\documentclass{article}

\usepackage{microtype}
\usepackage{graphicx}
\usepackage{subfigure}
\usepackage{booktabs} % for professional tables

\usepackage{hyperref}

\usepackage{bbm, dsfont}

\usepackage[accepted]{icml2025}

\usepackage{amsmath}
\usepackage{amssymb}
\usepackage{mathtools}
\usepackage{amsthm}

\usepackage[capitalize,noabbrev]{cleveref}

\theoremstyle{plain}
\newtheorem{theorem}{Theorem}[section]
\newtheorem{proposition}[theorem]{Proposition}
\newtheorem{lemma}[theorem]{Lemma}
\newtheorem{corollary}[theorem]{Corollary}
\theoremstyle{definition}
\newtheorem{definition}[theorem]{Definition}

\theoremstyle{remark}

\usepackage[textsize=tiny]{todonotes}

\DeclareMathOperator*{\argmax}{\arg\!\max}

\usepackage{enumerate}

\icmltitlerunning{Fixed-Confidence Multiple Change Point Identification}

\begin{document}

\twocolumn[
\icmltitle{Fixed-Confidence Multiple Change Point Identification \\
           under Bandit Feedback}

% It is OKAY to include author information, even for blind
% submissions: the style file will automatically remove it for you
% unless you've provided the [accepted] option to the icml2025
% package.

% List of affiliations: The first argument should be a (short)
% identifier you will use later to specify author affiliations
% Academic affiliations should list Department, University, City, Region, Country
% Industry affiliations should list Company, City, Region, Country

% You can specify symbols, otherwise they are numbered in order.
% Ideally, you should not use this facility. Affiliations will be numbered
% in order of appearance and this is the preferred way.
\icmlsetsymbol{equal}{*}

\begin{icmlauthorlist}
\icmlauthor{Joseph Lazzaro}{yyy}
\icmlauthor{Ciara Pike-Burke}{yyy}
%\icmlauthor{}{sch}
\end{icmlauthorlist}

\icmlaffiliation{yyy}{Department of Mathematics, Imperial College London, London, England}
% \icmlaffiliation{comp}{Company Name, Location, Country}
% \icmlaffiliation{sch}{School of ZZZ, Institute of WWW, Location, Country}

\icmlcorrespondingauthor{Joseph Lazzaro}{joseph.lazzaro18@imperial.ac.uk}

% You may provide any keywords that you
% find helpful for describing your paper; these are used to populate
% the "keywords" metadata in the PDF but will not be shown in the document
\icmlkeywords{Machine Learning, ICML}

\vskip 0.3in
]

% this must go after the closing bracket ] following \twocolumn[ ...

% This command actually creates the footnote in the first column
% listing the affiliations and the copyright notice.
% The command takes one argument, which is text to display at the start of the footnote.
% The \icmlEqualContribution command is standard text for equal contribution.
% Remove it (just {}) if you do not need this facility.

%\printAffiliationsAndNotice{}  % leave blank if no need to mention equal contribution
\printAffiliationsAndNotice{} % otherwise use the standard text.

\begin{abstract}

Piecewise constant functions describe a variety of real-world phenomena in domains ranging from chemistry to manufacturing. 
%where
In practice, 
it is often required to confidently identify the locations of the abrupt changes in these functions as quickly as possible. For this, we introduce a fixed-confidence piecewise constant bandit problem. Here, we sequentially query points in the domain and receive noisy evaluations of the function under bandit feedback. We provide instance-dependent lower bounds for the complexity of change point identification in this problem. These lower bounds illustrate that an optimal method should focus its sampling efforts adjacent to each of the change points, and the number of samples around each change point should be inversely proportional to the magnitude of the change. %with proportions 
%inverse to their magnitude. 
%inverse to the magnitude of each change. 
Building on this, we devise a simple and computationally efficient variant of Track-and-Stop and prove that it is asymptotically optimal in many regimes. We support our theoretical findings with experimental results in synthetic environments demonstrating the 
%computational 
efficiency of our method.

%\textbf{Keywords for openreview:} bandits, pure exploration, fixed confidence, change points

\end{abstract}

%%%%%%%%%%%%%%%%%%%%%%%%%%%%%%%%%%%%%%%%%%%%%%%%%%%%%%%%%%%%%%%%%%%%%%%%%%%%%%%
%%%%%%%%%%%%%%%%%%%%%%%%%%%%%%%%%%%%%%%%%%%%%%%%%%%%%%%%%%%%%%%%%%%%%%%%%%%%%%%
% INTRODUCTION
%%%%%%%%%%%%%%%%%%%%%%%%%%%%%%%%%%%%%%%%%%%%%%%%%%%%%%%%%%%%%%%%%%%%%%%%%%%%%%%
%%%%%%%%%%%%%%%%%%%%%%%%%%%%%%%%%%%%%%%%%%%%%%%%%%%%%%%%%%%%%%%%%%%%%%%%%%%%%%%
\section{Introduction} \label{section:intro}
% \begin{itemize}
%     \item Explain broadly what the problem is and some specific problem instances/applications in which this is useful
%     \item Briefly connect with existing work and explain why our particular problem is interesting and is not covered by existing work: in particular we study multiple change point identification unlike previous theoretical works which focused on single change point identification. We are also in the fixed confidence setting, unlike previous works.
%     \item List of contributions made in the paper
% \end{itemize}

% There are many settings in which a learner may be interested in identifying abrupt changes in mean rewards across an action space. For example in material development we may want to find the experimental conditions under which the behavior of a new material changes abruptly between phases \citep{park2021sequentialAdaptiveDesignForJumpRegression,ActiveLearningPiecewiseGP}; in Oceanology we may wish to map out the edge of a cliff on the seafloor \citep{ACPD}; or in communication and computer system engineering where we may want to study under what stress levels the system becomes overloaded and customer waiting times abruptly increase \citep{Active_Lan_2009}. However, data may be expensive to collect due to the cost, time, or compute of each experiment and hence it is important to study and construct sequential methods which can learn about the locations of these change points efficiently.

There are a variety of settings where it is necessary to confidently identify the location of change points in a piecewise constant function via sequential queries. For example, in communication and computer system engineering, 
the system can become overloaded and we need to determine the point at which this happens
%where we may want to study under what stress levels the system becomes overloaded and customer waiting times abruptly increase 
\citep{Active_Lan_2009}; 
%it is crucial to identify what temperatures (or chemical concentrations) particular crops or foods spoil to mitigate any health risks \textcolor{blue}{(need ref)}; 
during material development it is essential to find the experimental conditions under which the behavior of a new material changes abruptly between phases \citep{park2021sequentialAdaptiveDesignForJumpRegression,ActiveLearningPiecewiseGP}; 
%in aeronautical engineering we may want to stress test a wing design to see at what levels of wind cause catastrophic failure \textcolor{blue}{(need ref)}; 
%in energy fusion we may want to identify what range of plasma pressures recover a stable nuclear reaction \citep{fusion_paper}; 
or in Oceanology we may wish to map out the edge of a cliff on the seafloor \citep{ACPD}.
In all these settings, data is expensive to collect due to the cost, time, or compute of each experiment.
Moreover, there are often health or safety constraints which mean that we need to have high confidence in our results.
This motivates us to develop sequential methods which can learn about the locations of these change points efficiently and with high confidence. 
%Furthermore, in settings where the identified changes are important for health or safety it is important that we continue sampling until we are sufficiently confident in our results.
%

\begin{figure}[t]
\vskip -0.0in
\begin{center}
\centerline{\includegraphics[width=\columnwidth]{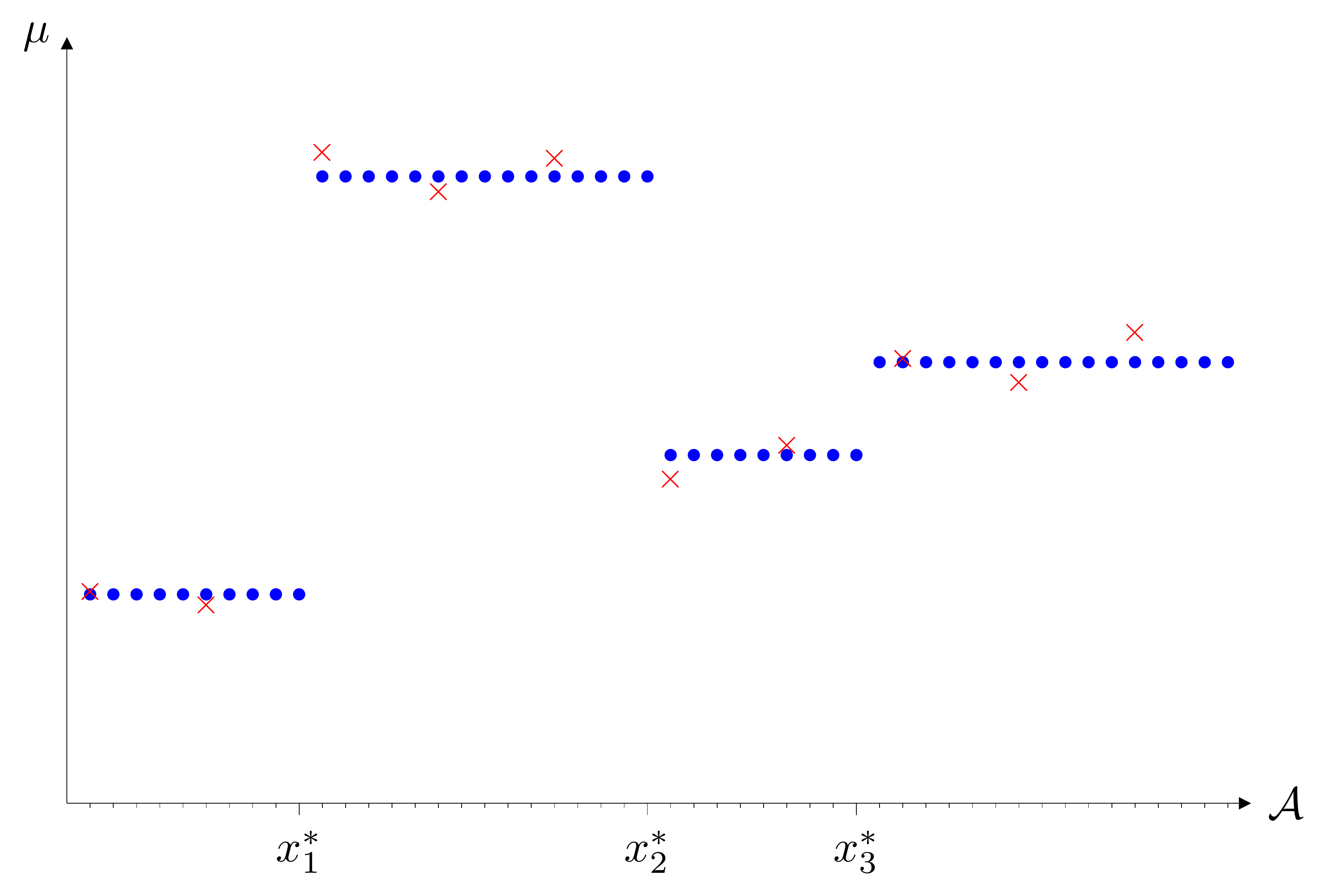}}
\caption{Example of an environment with piecewise constant expected rewards $\mu$ across the action space $\mathcal{A}$. Here the expected rewards for different actions in the action space are represented by blue dots. There are three change points in this environment at $x^*_1, x^*_2,$ and $x^*_3$. We additionally plot noisy rewards obtained from 10 arbitrary actions that have been played in $\mathcal{A}$ and represent these with red crosses.}
\label{fig_example_piecewise_constant}
\end{center}
\vskip -0.4in
\end{figure}

In this paper we consider a multi-armed bandit setting in which the expected rewards are piecewise constant across an action space $\mathcal{A}$, see Figure \ref{fig_example_piecewise_constant} for an example. % \footnote{Note this is distinct from non-stationary settings where the mean rewards are piecewise-constant over \emph{time}.
%rather than over $\mathcal{A}$. See Section \ref{section:related_work} for discussion.}. 
In each round $t=1,2,\dots$ we select an action in our action space and observe the 
%piecewise constant 
mean reward at that action plus some random noise (see Section \ref{section:problem_setting} for more details).
We assume the mean reward function is stationary across time, but piecewise constant across $\mathcal{A}$.
The objective of the learner is to confidently identify the locations of the $N$ change points/jumps in the mean reward function with the fewest number of samples possible.
While there exist methods for locating the minimum of a smooth or convex function under bandit feedback \citep[e.g.][]{ZoomingAlgo,GPUCB_paper,treeAlgo}, the abrupt changes in the mean reward function across the action space make these methods inapplicable to our setting.
%There has been work on change point identification with bandit feedback under a fixed-budget constraint and assuming that there is a single change point 
There has been work on identification of a single change point across the action space with bandit feedback, under a fixed-budget constraint
\citep{active_Hall2003,Active_Lan_2009,lazzaro2025fixedbudgetchangepointidentification}. However, in practice it is often necessary to continue sampling until we are sufficiently confident we have identified $N\geq1$ change points correctly, at which point we stop playing.
For example, in material development we need to identify the phase transitions of an unfamiliar new material with high confidence for the sake of both production and safety precautions \citep{park2021sequentialAdaptiveDesignForJumpRegression,ActiveLearningPiecewiseGP}. It is therefore important to study this fixed-confidence variant of the problem, which we will refer to as the fixed-confidence piecewise constant bandits problem. 

In this paper we provide a comprehensive study of the fixed-confidence piecewise constant bandits problem. 
%We consider different assumptions on how much information the learner has about the number of change points, as it turns out this will determine the complexity of the problem. %an we illustrate that its difficulty will depend on how much information is given to the learner.
%Now, analysis of this problem will depend on how much information is given to the learner. 
We firstly study the complexity of searching for $N$ change points, when the true number of change points is known to be exactly $N$ (see Definition \ref{defn:multiple_exact}). However, in practice, there may be an unknown number of additional changes present and it is important to develop methods robust to this. For example, when looking for a cliff on the sea floor, there may be other smaller changes in the sea-depth elsewhere. Hence, we also study the complexity of finding any $N$ change points out of an unknown number of changes $m \geq N$ (see Definition \ref{defn:multiple_any}). In both cases we prove a lower bound on the expected sample complexity (Theorems \ref{theorem_exact_m_lb}, \ref{theorem_any_m_lb_general_case}), which provides some key insights. Firstly, it tells us that an optimal method should focus most of its sampling efforts on actions immediately adjacent to each of the change points. Secondly, the number of samples adjacent to each change should be inversely proportional to its magnitude. We use these insights to construct the Multiple Change Point Identification (MCPI) policy (Algorithm \ref{algo_TandS2}), a variant of Track-and-Stop \citep{kaufmann_16a}, which we show to be asymptotically optimal (Theorem \ref{theorem_upper_mcp_asymptotic}, for both objectives). Unlike some other variants of Track-and-Stop \citep{kaufmann_16a,Juneja2018SampleCO_TAS_GENERALISATION}, MCPI is simple and computationally inexpensive. In Section \ref{section:experiments} we conduct experiments in synthetic environments to illustrate MCPI is optimal and that it outperforms existing related works in Clustering Bandits \citep{Yang2022OptimalClustering}.

%%%%%%%%%%%%%%%%%%%%%%%%%%%%%%%%%%%%%%%%%%%%%%%%%%%%%%%%%%%%%%%%%%%%%%%%%%%%%%%
%%%%%%%%%%%%%%%%%%%%%%%%%%%%%%%%%%%%%%%%%%%%%%%%%%%%%%%%%%%%%%%%%%%%%%%%%%%%%%%
% RELATED WORK
%%%%%%%%%%%%%%%%%%%%%%%%%%%%%%%%%%%%%%%%%%%%%%%%%%%%%%%%%%%%%%%%%%%%%%%%%%%%%%%
%%%%%%%%%%%%%%%%%%%%%%%%%%%%%%%%%%%%%%%%%%%%%%%%%%%%%%%%%%%%%%%%%%%%%%%%%%%%%%%
\section{Related Work} \label{section:related_work}
% \begin{itemize}
%     \item Fixed confidence pure exploration: there has been minimax works on a variety of problems however they are oftentimes asymptotically suboptimal; there has also been recent works on asymptotcially optimal methods in a variety of settings. Explain that we will be adapting one of these settings (track and stop) and this will be our goal in the paper which we will discuss later. Could also mention here that relevant work could be clustering bandits and mention the minimax and asymptotic versions here.
%     \item Change points in non-stationary bandits: Explain that there are change points in non-stationary bandits but this is different from our setting.
%     \item Other change point identification bandits works: explain that they are anytime with no theory or are fixed budget. Explain that the theoretical ones also focus on single change point/curve problems (need to check the castro L2 optimal one... to confirm this)
% \end{itemize}

%\paragraph{Fixed Confidence Pure Exploration} 
\textbf{Fixed Confidence Pure Exploration}\quad The most well-studied fixed-confidence pure exploration problem is best-arm identification where the learner aims to identify the action with the largest mean reward. While some works have focused on minimax optimal sample complexity \citep[see][]{fixed_confidence_survey}, these methods are often instance-dependent asymptotically suboptimal. More recent works have focused asymptotic optimality \citep[e.g. Track-and-Stop and Top-Two methods by][resp.]{kaufmann_16a,TopTwoJourdan}. Our method is a variant of the Track-and-Stop approach \cite{kaufmann_16a}.
Normally, when using Track-and-stop methods, the actions played are chosen after solving an optimisation problem in each round. As these optimisation problems often have to be solved numerically, the choice of action in each round can be somewhat unintuitive and potentially computationally expensive. However, in the proposed change point identification setting, we can explicitly solve the corresponding optimisation problems to provide a computationally efficient and intuitive variant of Track-and-Stop. This is discussed in more detail in Section \ref{section:one_cp}.
%%Although, as we show later, the structure of our problem setting allows us to develop a more interpretable and computationally inexpensive variant of Track-and-Stop.
%(Gairiver & Kaufmann, 2016) is a seminal work and their tracking ideas have motivated the class of track-and-stop strategies and theory for a wide variety of works, including our own. This is because the lower bounds they construct for best arm identification are also applicable to a wide variety of other pure exploration bandits problems (Ch33.2, Lattimore & Sepesvari, 2020). However, the bounds they present are in implicit form (i.e. as the solution of an optimisation problem). Hence, their tracking methods are computationally expensive and somewhat unintuitive. In our work we are able to build off the bounds from (Garivier & Kaufmann, 2016) to provide intuitive instance-dependent lower bounds which can then, in turn, be used to define computationally efficient and intuitive tracking procedures for our setting.

% method can be  by solving an optimization problem explicitly in our setting, making our method more interpretable and computationally inexpensive.

There has also been recent work on fixed confidence Clustering Bandits \citep{Yang2022OptimalClustering,thuot2024activeclusteringbanditfeedback,yavas2024generalframeworkclusteringdistribution}, where there are exactly $N$ distinct mean rewards 
%arbitrarily located in the environment. 
which any arm in the action space can have.
The objective is to partition the action space into $N$ sets of arms, such that each set contains arms with the same mean reward. Unlike the problem presented in this paper, in the clustering bandits problem there is no (piecewise constant) structure of the mean rewards across the action space. Hence these methods are significantly suboptimal when applied to the fixed-confidence piecewise constant bandit problem. Furthermore
%, unlike the work in Clustering Bandits, 
we additionally consider settings where we do not know the number of change points/clusters $N$. We discuss this further in Section \ref{section:experiments}.

%\paragraph{Change points \& Non-Stationary Bandits} 
% There is a related but distinct literature on bandit problems with non-stationary environments that change over time, oftentimes focused on regret minimization \citep[e.g.][]{Discounted_UCB_non_stationary,non_stationary_CP_auer, non_stationary_CP_chen, banditsQuickestCDP}.  These settings have made use of the change point analysis literature in order to detect when the mean rewards have changed over time \citep[for a survey see][]{Basseville1993DetectionOA,CP_survey}. In contrast to these works, in our setting the environment is stationary and does not change over time, but instead the mean rewards change abruptly \emph{across the action space}. 
\textbf{Change points \& Non-Stationary Bandits}\quad
There is a vast literature in statistics devoted to the detection of change points in a given data set \citep[for survey see][]{CP_survey}. 
Change points have appeared in the bandit literature in the context of non-stationary environments where the expected rewards from the actions change \emph{over time} \citep[e.g.][]{Discounted_UCB_non_stationary,non_stationary_CP_auer, non_stationary_CP_chen, banditsQuickestCDP, hou2024almost}. We emphasize that this is distinct from the setting considered in this paper where the environment is stationary across time with abrupt changes in the expected rewards of the actions \emph{across the action space}.

%\paragraph{Change Points Across The Action Spaces} 
% Multiple anytime methods have been proposed for settings with abrupt changes in the mean reward function across the action space. For example, \citet{AdaptiveDesignSupercomputer,park2021sequentialAdaptiveDesignForJumpRegression,ActiveLearningPiecewiseGP} propose Bayesian Optimization methods for actively approximating an entire piecewise-smooth mean reward function under bandit feedback. Similarly, \citet{ACPD} also provide a method to identify change points in a mean reward function, given an inputted model for the type of change point of interest. However these methods have only been assessed empirically and are not accompanied by theoretical analysis, whereas in this paper we will focus on fixed confidence versions of the problem and complementary theoretical analysis and optimality for our methods.
\textbf{Change Points Across The Action Spaces}\quad
Anytime methods have been proposed to identify change points or actively learn an entire piecewise smooth function under bandit feedback \citep{AdaptiveDesignSupercomputer,park2021sequentialAdaptiveDesignForJumpRegression,ActiveLearningPiecewiseGP, ACPD}. These methods use variants of Bayesian Optimization, but have only been assessed empirically and are not accompanied by theoretical analysis. In this paper we focus on deriving computational efficient policies that are also theoretically optimal.

% Under a fixed budget assumption on the total number of samples the learner can make $T$, multiple other methods have also been proposed. \citet{Active_Castro2005} studied (instance-independent) near-minimax two-stage method for active learning of a whole piecewise constant function in $\mathbb{R}^d$. Similarly, under fixed budget assumptions, single change point (or change surface) identification has also been studied by \citet{Active_Lan_2009,active_Hall2003} who provided multistage-methods with asymptotic guarantees (as $T \rightarrow \infty$). Moreover, \citet{lazzaro2025fixedbudgetchangepointidentification}  provide non-asymptotic minimax optimal binary search methods for identifying a single change point. Our objective, however, will not be focused on a fixed budget on the number of samples, but instead we consider a fixed confidence that our final estimates for the multiple change points are correct. Furthermore, we will discuss how to deal with settings in which there is incomplete knowledge on the true number of change points in the mean rewards.

Other works have also studied settings with change points across the action space under a fixed budget assumption,  where the total number of samples the learner can make 
%to locate the change points 
is bounded by a fixed $T>0$. For example, \citet{Active_Castro2005} proposed an (instance-independent) nearly minimax optimal two-stage method for active learning of a whole piecewise constant function in $\mathbb{R}^d$. Similarly, \citet{Active_Lan_2009,active_Hall2003} provided multistage-methods with asymptotic guarantees (as $T \rightarrow \infty$) for single change point (or change surface) identification in the fixed budget setting. Recently, \citet{lazzaro2025fixedbudgetchangepointidentification} provided non-asymptotic minimax optimal binary search methods for identifying a single change point in the fixed budget setting.
In contrast, we consider \emph{fixed-confidence} methods for identifying change points. 
Moreover, most prior work has focused 
%As well as considering the fixed budget setting, the existing work on change point identification under bandit feedback has also focused 
on settings with exactly one change point (or surface), while we consider identifying $N\geq 1$ change points. %In this paper we consider settings with $N$ change points and focus on \emph{fixed confidence} identification of the change points.
%. Furthermore, although it is interesting to consider a limited budget on the number of samples, in practice it may be more important for us to only stop when we are sufficiently confident in our estimates for the change points. However, this fixed confidence version of change point identification has not been studied in the literature and this is a gap we intent to fill in this paper. 

We also point out the concurrent work by \citet{bacchiocchi2025regret}, which studies a regret minimization variant of our problem setting in which the learner tries to maximize cumulative rewards received when the underlying mean rewards are piecewise linear across the action space. In this paper, however, we specifically focus on optimally identifying the location of the change points under a fixed-confidence assumption.

%%%%%%%%%%%%%%%%%%%%%%%%%%%%%%%%%%%%%%%%%%%%%%%%%%%%%%%%%%%%%%%%%%%%%%%%%%%%%%%
%%%%%%%%%%%%%%%%%%%%%%%%%%%%%%%%%%%%%%%%%%%%%%%%%%%%%%%%%%%%%%%%%%%%%%%%%%%%%%%
% PROBLEM SETTING
%%%%%%%%%%%%%%%%%%%%%%%%%%%%%%%%%%%%%%%%%%%%%%%%%%%%%%%%%%%%%%%%%%%%%%%%%%%%%%%
%%%%%%%%%%%%%%%%%%%%%%%%%%%%%%%%%%%%%%%%%%%%%%%%%%%%%%%%%%%%%%%%%%%%%%%%%%%%%%%
\section{Problem Setting} \label{section:problem_setting}
% \begin{itemize}
% \item explain the sequential aspect...
%     \item Explain more specifically what the form the environments can take, namely piecewise constant. (Maybe make the assumption here that the change points are at least 1 apart... and explain why this is reasonable.)
%     \item Assume noise is Gaussian with known variance sigma squared, put this in with the first point...
%     \item Explain how this fixed confidence setting works: we will have an objective (we will consider multiple throughout the paper). Then, given this objective, we will need to define three things to construct a method - a selection rule, a stopping rule, and a recommendation rule while guaranteeing our objective. We will then try to minimize the expected number of samples required for our method to attain our objective and stop.
% \end{itemize}

% \begin{itemize}
%     \item Need to define the actions/rewards/means notation here. Explain that we sequentially make decisions and that the noise is N(0,1)
% \end{itemize}

We consider a multi-armed bandit setting where there are $K$ arms ($\mathcal{A}=[K]$)\footnote{Here we denote $[K]=\{1,\dots,K\}$.}  with mean rewards $(\mu_i)_{i=1}^K$.
Note that this is equivalent to a discretization of the continuous action analogue of the problem, see \citep{lazzaro2025fixedbudgetchangepointidentification}. We assume that the mean rewards are piecewise constant. Let $\underline{x}^* \subset [K-1]$ be the set of $N$ change points with $|\underline{x}^*|=N$. 
We use the convention that if there is a change point at $i \in\underline{x}^*$, then the mean reward changes between the $i$th and $(i+1)$st arm, i.e. $\mu_i \neq \mu_{i+1}$. Conversely, for any arm that is not a change point, i.e. $i \in [K-1]\backslash \underline{x}^*$, we have $\mu_i = \mu_{i+1}$. % otherwise for any $i \in\underline{x}^*$ $\mu_i \neq \mu_{i+1}$.
In each round $t =1,2, \dots$ we play an action $a_t \in [K]$ and observe a reward $y_t = \mu_{a_t} + \epsilon_t$, where the noise $\epsilon_t \sim N(0, \sigma^2)$ is i.i.d. Gaussian with mean zero and variance $\sigma^2 \in \mathbb{R}_+$. We assume that the noise variance is known and $\sigma^2=1$ for simplicity. The assumption of a known variance (proxy) is common in the fixed-confidence pure exploration literature \citep[e.g.][]{chen_combinatorial,pmlr-v35-jamieson14_lilucb,GaussianBAI}.

We define $V_{K,N}$ as the set of environments with piecewise constant mean rewards
%$(\mu_i)_{i=1}^K$ 
when there is a total of $K$ actions and $N$ change points. In particular, for any $v \in V_{K,N}$ we denote the mean rewards in $v$ as $(\mu_{i,v})_{i=1}^K$ and the set of change points in $v$ as
\begin{equation*}
    \underline{x}^*_{v} = \{j \in [K-1]: \mu_{v,j} \neq \mu_{v,j+1}\}.
\end{equation*}
We denote the elements of the set of change points as
\begin{equation*}
    \underline{x}^*_{v} = \{x^*_{v,1}, \dots, x^*_{v,N}\}
\end{equation*}
such that  they are indexed from left to right, namely $x^*_{v,1} < \dots < x^*_{v,N}$. Furthermore, we assume that the change points are separated by at least one action, namely 
\begin{equation*}
    \forall i \in [K-1], \quad x^*_{v,i} + 1 < x^*_{v,i+1}.
\end{equation*}
This is reasonable when considering our finite action space as a discretization of a continuous one which is sufficiently fine. 
We use $v(i)$ to denote the reward distribution of arm $i$ in environment $v$, namely $N(\mu_{v,i},1)$, and
$D\left(\cdot,\cdot\right)$
%$D\left(v(i),v'(i)\right)$ 
to denote the KL-divergence between two distributions.
%to separate change points at intervals of interest. 
We additionally denote the size of the change in mean at the $j$th change point in environment $v$, $x^*_{v,j}$, as %(i.e. between $x^*_{v,j}$ and $x^*_{v,j}+1$) as 
\begin{equation*}
    \Delta_{v,j} = \big|\mu_{v,x^*_{v,j}} - \mu_{v,x^*_{v,j}+1}\big|
\end{equation*}
Finally, we denote $x^*_{v,(i)}$ as the change point with the $i$'th largest magnitude and denote its magnitude by $\Delta_{v,(i)}$. Hence, $\Delta_{v,(1)} \geq \dots \geq \Delta_{v,(N)}$.
We will drop the $v$ notation when it is clear which environment we are considering. 
%\textcolor{blue}{(maybe will include a tikz plot which has this version of the labelling to help clarify things...)} 

% Finally, note that we will also denote the reordering of the set of pairs of change points and respective size of change in means, 
% \begin{equation*}
%     \{(x^*_{v,1},\Delta_{v,1}), \dots, (x^*_{v,N},\Delta_{v,N})\},
% \end{equation*}
% such that they are ordered from largest change in mean to smallest as
% \begin{equation*}
%     \{(x^*_{v,(1)},\Delta_{v,(1)}), \dots, (x^*_{v,(N)},\Delta_{v,(N)})\},
% \end{equation*}
% where $\Delta_{v,(1)} \geq \dots \geq \Delta_{v,(N)}$.

In the fixed confidence setting we consider, the learner will be given a small fixed \textbf{confidence level} $\delta \in (0,1)$ and an \textbf{objective} for their change point estimates to be ``correct''. There are several natural objectives we can consider in this problem, as outlined in Sections \ref{section:one_cp} and \ref{section:multiple_cp}. The goal of the leaner is to satisfy this objective with probability greater than $1-\delta$, while minimizing the number of samples required to do so.
%We will formalize different objectives later.
%We will discuss/formalize different potential objectives in the following sections. 
%
In particular, we propose methods which sequentially select actions to play in each round until they are confident of the location of the change points, at which point they return the estimated change points.
%given the observed reward from actions played in previous rounds, and stop playing to return the estimated locations of the change points when we are confident that the objective has been satisfied.
More formally,
%
%For this, 
we define a \emph{policy} $\pi$ to consist of three rules:   
% a fixed confidence settings we must define the following three rules to satisfy our objective.
\begin{enumerate}
    \item \textbf{Sampling Rule:} A procedure to determine which action to play in each round, given the sequence of actions and rewards observed so far.
    \item \textbf{Stopping Rule:} A rule defining a stopping time $\tau$ at which point we determine we have collected sufficient data and the policy stops. 
    \item \textbf{Recommendation Rule:} A rule to return our final estimate for the set of change points. After stopping time $\tau$, we denote the estimate as $\underline{\hat{x}}_\tau$.
\end{enumerate}
A good policy will return an estimate $\underline{\hat{x}}_\tau$ that satisfies our objective with a small expected stopping time (i.e. sample complexity), $\mathbb{E}_{\pi,v}[\tau]$, and stops almost surely in finite time, $\mathbb{P}_{\pi,v}(\tau<\infty)=1$. Here, we define $\mathbb{P}_{\pi,v}$ to be the measure induced by all interactions between a policy $\pi$ and an environment $v$, dropping the subscripts when it is clear which policy and environment we refer to.
%%%%%%%%%%%%%%%%%%%%%%%%%%%%%%%%%%%%%%%%%%%%%%%%%%%%%%%%%%%%%%%%%%%%%%%%%%%%%%%
%%%%%%%%%%%%%%%%%%%%%%%%%%%%%%%%%%%%%%%%%%%%%%%%%%%%%%%%%%%%%%%%%%%%%%%%%%%%%%%
% WARM UP
%%%%%%%%%%%%%%%%%%%%%%%%%%%%%%%%%%%%%%%%%%%%%%%%%%%%%%%%%%%%%%%%%%%%%%%%%%%%%%%
%%%%%%%%%%%%%%%%%%%%%%%%%%%%%%%%%%%%%%%%%%%%%%%%%%%%%%%%%%%%%%%%%%%%%%%%%%%%%%%
\section{Warm-Up: Exactly One Change Point} \label{section:one_cp}
% Although our aim is to develop methods to confidently identify $N$ change points, we begin by considering the case where there is a single change point. This allows us to present the key insights that we will build upon in the multiple change point setting in Section \ref{section:multiple_cp}. Note that even when there is a single change point, there are two possibilities; either the learner knows there is exactly one change point or there could be multiple change points but we just want to detect one. In Section \ref{section:multiple_cp}, we will show that a single algorithm can be optimal for both possibilities. For now, however, we focus on the former case where the learner knows there is exactly one change point.

Although our aim is to develop methods to confidently identify $N$ change points, we begin by considering the case where there is a single change point. This allows us to present the key insights that we will build upon in the multiple change point setting in Section \ref{section:multiple_cp}. In this section, we additionally assume that the learner knows there is exactly one change point.
%In Section \ref{section:multiple_cp}, we will show that a single algorithm can be optimal in this setting as well when there are multiple change points, but we just want to identify one. 
%In Section \ref{section:multiple_cp}, we will show that a single algorithm can be optimal for identifying one change point in this setting where we restrict the possible number of change points to be exactly one as well as for settings in which there are potentially additional change points present across the action space.
In Section \ref{section:multiple_cp}, we will show that a single algorithm can be optimal in this setting as well when as there are potentially multiple change points, but we just want to identify one. 

%In order to help motivate and explain our approach to the multiple change point setting, we first simplify and consider the problem of identifying only one change point.

\subsection{Instance Dependent Lower Bound}
% \begin{itemize}
%     \item Explain the objective of identifying one change point given that we know that there is exactly one change point (maybe write in a definition and give motivations/applications)
%     \item Give lower bound which has the inf displayed
%     \item Give lower bound which has the inf solution
%     \item Explain the track-and-stop intuition that comes from this
%     \item Mention that we will provide a method which can attain this asymptotically below.
%     \item emphasize that we are able to explicitly solve the inf...
% \end{itemize}

Suppose that we are in an environment with exactly one change point and the learner is given this information. Namely, it is known that $|\underline{x}^*|=1$. In this simple setting, we want our policy to return an estimate for the change point (after stopping) which is equal to the true change point with probability greater than $1-\delta$. This should happen after a finite number of samples almost surely. We refer to a policy which satisfies this objective as an Exact-$(1,\delta)$ policy.

\begin{definition}\label{defn_exact1_objective}
    \textbf{Exact-$(1,\delta)$ Policy}: For any $v \in V_{K,1}$, an Exact-$(1,\delta)$ policy $\pi$ with stopping time $\tau$ returns an estimate $\hat{x}_\tau \in [K-1]$ satisfying 
\begin{align}
    \mathbb{P}_{\pi,v} (\hat{x}_\tau = x^*_{v,1}) &> 1-\delta, \label{eqn_1cp_defn1}\\
    \mathbb{P}_{\pi,v} (\tau < \infty) &=1. \label{eqn_1cp_defn2}
\end{align}
\end{definition}

For any Exact-$(1,\delta)$ policy, we provide the following non-explicit, instance-dependent lower bound on its expected stopping time, which is a modification of the general lower bound first described by \citet{kaufmann_16a}. See Appendix \ref{app_proof_theorem_1cp_general} for the proof.

\begin{theorem} \label{theorem_1cp_general}
    For any Exact-$(1,\delta)$ policy $\pi$ with stopping time $\tau$ 
    %the expected stopping time 
    in environment $v \in V_{K,1}$ we have
    %Suppose that a policy $\pi$ with stopping time $\tau$ is Exact-$(1,\delta)$. Then a lower bound for the expected stopping time in environment $v \in V_{K,1}$ is 
    \begin{equation*}
        \mathbb{E}_{\pi,v}[\tau] \geq c^*(v) \log\left(\frac{1}{4\delta}\right),
    \end{equation*}
    where we define
    \begin{equation} \label{eqn_1cp_opt_problem}
        c^*(v)^{-1} =  \sup_{\alpha \in \mathcal{P}_{K}}\inf_{v' \in V^{alt}_{K,1}(x^*_v)} \sum_{i=1}^K \alpha_i D\left(v(i),v'(i)\right).
    \end{equation}
    Here $V^{alt}_{K,1}(x^*_v)$ denotes the set of environments in $V_{K,1}$ which have change point not equal to $x^*_v$ and $\mathcal{P}_{K}$ is the standard $K$-dimensional simplex.
\end{theorem}

From \citep{kaufmann_16a} we know that solving the optimization problem in \eqref{eqn_1cp_opt_problem} can be informative both for understanding the complexity of the problem as well as for understanding what an optimal strategy would be. Let $\alpha_i^*(v)$ be the weights which solve the optimization problem \eqref{eqn_1cp_opt_problem}, namely
\begin{equation*}\label{alpha_star_opt_problem_1cp}
    c^*(v)^{-1} =  \inf_{v' \in V^{alt}_{K,1}(x^*_v)}\sum_{i=1}^k \alpha_i^*(v) D(v_i,v_i').
\end{equation*} 
These $\alpha_i^*(v)$ determine the optimal proportion of samples we should allocate to each action, $i \in [K]$, in environment $\nu$. 
While this optimization problem \eqref{eqn_1cp_opt_problem} is typically expensive to solve \citep{Degenne2019NonAsymptoticPE}, we are able to solve \eqref{eqn_1cp_opt_problem} explicitly to show the unique optimal weights 
%for each action 
are
% Unlike many other works we are able to explicitly solve this optimization problem \eqref{eqn_1cp_opt_problem} to show the unique optimal weights for each action are
    \begin{equation} \label{eqn_1cp_alpha_star}
   \alpha_i^*(v)=
        \begin{cases}
    1/2& \text{if } i=x^*_v \text{\,or\,} x^*_v+1\\
    0             & \text{otherwise.}
\end{cases}
    \end{equation}
    This leads to
\begin{equation*}
    c^*(v)^{-1} = \Delta_1^2/8\sigma^2.
\end{equation*} 
%Whereas normally this optimization problem would be expensive to solve \citep{Degenne2019NonAsymptoticPE}. 
This proves Corollary \ref{theorem_exact_1cp_lb_explicit} below.
This suggests that an optimal Exact-$(1,\delta)$ policy will asymptotically focus all of the samples immediately either side of the 
%change in mean reward in our environment
change point, and will play these two actions equally often. This observation motivates the algorithms we develop in subsequent sections. See Appendix \ref{app_proof_theorem_exact_1cp_lb_explicit} for the proof of Corollary \ref{theorem_exact_1cp_lb_explicit}.

\begin{corollary}\label{theorem_exact_1cp_lb_explicit}
For any Exact-$(1,\delta)$ policy $\pi$ with stopping time $\tau$,
    %Suppose that a policy $\pi$ with stopping time $\tau$ satisfies the Exact-$(1,\delta)$ Objective. Then 
    a lower bound for the expected stopping time in environment $v \in V_{K,1}$ with a single change in mean of magnitude $\Delta$ is
    \begin{equation*}
        \mathbb{E}_{\pi,v}[\tau] \geq \frac{8\sigma^2}{\Delta^2} \log\left(\frac{1}{4\delta}\right).
    \end{equation*}
\end{corollary}

\subsection{Track-and-Stop Approach and Upper Bounds}
% \begin{itemize}
%     \item Go through three main things of sampling rule, stopping rule, and recommendation rule - giving intuition
%     \item Mention the reasons we have chosen to do track and stop instead of top two here as well! (definitely mention somewhere above if not)
%     \item Mention algorithm will satisfy both objectives
%     \item Formally write algorithm here
%     \item Theorem: non-asymptotic upper bound
%     \item Theorem: asymptotic upper bound and comparison with lower bounds
% \end{itemize}

% \begin{itemize}
%     \item Mention here or earlier that we will assume knowledge of the vaiance $\sigma =1$
% \end{itemize}

In other fixed confidence bandits settings, there are two popular methods to attain asymptotic optimality. The first are Track-and-Stop methods \citep[e.g.][]{kaufmann_16a,Degenne2019NonAsymptoticPE,Degenne2019PureEMultipleCorrectAnswers,Juneja2018SampleCO_TAS_GENERALISATION} which focus on first developing lower bounds and then playing actions according to approximations of the optimal proportions $\alpha^*$ (which appear in the lower bounds). Secondly, in Top-Two methods \citep[e.g.][]{TopTwoRusso,Qin_TopTwoEI,TopTwoJourdan} the learner alternates between playing a `leader' action and a `challenger' action which represent potential positions of the target/objective in the action space. Since we have shown in Corollary~\ref{theorem_exact_1cp_lb_explicit} that for our problem the optimal proportions $\alpha^*$ have a simple closed form expression, it is natural to consider Track-and-Stop approaches rather than Top-Two methods. Furthermore, Track-and-Stop methods have been studied in settings with multiple correct answers \citep{Degenne2019PureEMultipleCorrectAnswers, chen2025optimalmultiobjectivebestarm}, 
%which recommend a set of actions after stopping, 
leading us to believe they will be more suited to the multiple change point setting than Top-Two methods which have only been studied in settings where it is required to return one action. Finally, the main downside to Track-and-Stop methods in other settings is their potentially very large computational complexity. This is not an issue in our setting since we can explicitly solve the optimization problem in Corollary \ref{theorem_exact_1cp_lb_explicit}. 
%Whereas in other settings this optimization problem would need to be solved numerically at every time step. 
We now show how to adapt Track-and-Stop methods, using Corollary \ref{theorem_exact_1cp_lb_explicit}, to construct our policy called Single Change Point Identification (CPI) which is Exact-$(1,\delta)$ and has optimal sample complexity. CPI is stated explicitly in Algorithm \ref{algo_TandS}.

\subsubsection{Sampling Rule}
Suppose we knew we were in environment $v$. Then from our lower bound in Corollary \ref{theorem_exact_1cp_lb_explicit} and equation \eqref{eqn_1cp_alpha_star}, we would optimally play half of our actions adjacently either side of the change point $x^*_v$ (i.e., we would play actions $x^*_v $ and $ x^*_v+1$ each half of the time and not play any other actions). However, in practice we do not know the environment $v$ and therefore we do not know which actions $x^*_v$ and $x^*_v+1$ to play. Hence, we propose the following tracking and forced exploration rules, which are of the general form of D-Tracking introduced by \citet{kaufmann_16a} for best-arm identification. We note that there are alternative tracking rules in the Track-and-Stop literature such as C-tracking and proportional tracking, however these have been shown to be less effective empirically \citep{kaufmann_16a,Degenne2019NonAsymptoticPE}.

\textbf{Tracking}\quad For the tracking component, we first estimate the change point in each round $t$, $\hat{x}_t$.
%, and play an action adjacent either side of this estimate.
We propose to use the following simple estimator for the change point which selects the action corresponding to the largest empirical change in mean, namely
    \begin{equation}\label{eqn_estimator_simple_defn}
        \hat{x}_t(S) = \text{argmax}_{a \in S} |\hat{\mu}_a(t) - \hat{\mu}_{a+1}(t)|.
    \end{equation}
 % with\footnote{$S$ will later denote the set of potential remaining change points.} $S = [K-1]$ and $\hat{\mu}_a(t)$ is the empirical mean of the rewards obtained from playing action $a$ up to time $t$.
Here $S$ is the set of potential change points, which in this case is $[K-1]$, but may change when we consider multiple change points in Section \ref{section:multiple_cp}. Here $\hat{\mu}_a(t)$ is the empirical mean of the rewards obtained from playing action $a$ up to time $t$.
 We choose this estimator due to its simplicity, but we expect alternative estimators, such as least squares, could also lead to similar results. %For now, however it is simpler to consider $\hat{x}_t$ defined in \eqref{eqn_estimator_simple_defn}. In the future it would be interesting to consider different types of estimators and how these might improve both non-asymptotic theoretical results as well as empirical performance of proposed policies.\\

Once we have an estimate for the change point in round $t$, $\hat{x}_t$, we can establish the tracking criteria by noting that if $\hat{x}_t$ were the true change point then we would want to play equally either side of this change in mean. 
%then the optimally played proportions in an environment with change point $\hat{x}_t$ is half on either side of this estimate. 
Consequently, we estimate the optimal proportions for each action at time $t$ as $\hat{\alpha}_i(t)=1/2$ if $i \in \{\hat{x}_t,\hat{x}_t+1\}$, and $0$ otherwise. Using these estimates, our tracking criteria is to play the action with the proportion of plays furthest below the estimated optimal proportions in round $t$. Let $T_i(t)$ be the total number of times we have played action $i$ up to round $t$. Then in round $t$, we play
\begin{equation} \label{eqn_tracking}
    a_t = \text{argmin}_{i\in \{1,\dots, K\}} \left(\frac{T_i(t)}{t} - \hat{\alpha}_i(t)\right).
\end{equation}
This is equivalent to playing whichever arm has been played least out of $\{\hat{x}_t,\hat{x}_t+1\}$, namely
% \begin{equation*}
%     a_t = \text{argmin}_{i \in \{\hat{x}_t,\hat{x}_t+1\}} T_i(t),
% \end{equation*}
$a_t = \text{argmin}_{i \in \{\hat{x}_t,\hat{x}_t+1\}} T_i(t)$,
as in Algorithm \ref{algo_TandS} line 8.

%\paragraph{Forced Exploration}
\textbf{Forced Exploration}\quad If we only do the above tracking, the quality of our estimates $\hat{x}_t$ may improve slowly over time (if at all). Hence we additionally include some forced exploration across the action space, similar to previous works \citep[e.g.][]{garivier2018thresholdingbanditdoserangingimpact,Yang2022OptimalClustering}. For this, if at any time $t$, we have an action $i \in [K]$ that has been played less than $\sqrt{t}$ times, then we play that action (breaking ties arbitrarily) as seen in Algorithm \ref{algo_TandS} lines 5-6. The idea is that this forced exploration is large enough to force our estimates for the change point and optimal proportions to be close to their true values of $x^*$ and $\alpha^*$. Simultaneously, this forced exploration should not be so large that our policy becomes overly exploratory and suboptimal. 
%Choosing approximately $\sqrt{t}$ for our forced exploration has worked well in other settings (\textcolor{blue}{reference other settings/papers}).
We will show in Theorem \ref{TAS_asymp_UB} that this combination of tracking and forced exploration is sufficient to achieve asymptotic optimality.

% \paragraph{Edge Case (Multiple Equally Sized Changes)} \textcolor{blue}{Move this to the next section} In the case where there are multiple equally sized changes, for example $\Delta_{v,(1)}=\Delta_{v,(2)}$, we see from Theorem \ref{theorem_any_1cp_lb_explicit} that the lower bound only includes the complexity from finding the position of any single \emph{one} of the largest changes. Hence, an optimal policy would simply focus its sampling efforts around any \emph{one} of the largest changes. To avoid going back and forth sampling near different changes of equal size, we include a condition that we only update our estimate $\hat{x}_t$ when there is evidence of a change in mean significantly larger than anywhere else. In particular this is when there exists $i \neq j \in [K-1]$ such that 
% \begin{equation}\label{eqn_edge_case_condition}
%     |\hat{\mu}_i(t) - \hat{\mu}_{i+1}(t)| > |\hat{\mu}_j(t) - \hat{\mu}_{j+1}(t)| + r(t)
% \end{equation}
% holds with $r(t)$ defined as
% \begin{equation}\label{eqn_radius_definition}
%     r(t) = \sqrt{\frac{4\log(t) + 2\log(2\log(t)) + 1/2}{t}}.
% \end{equation}
% Note however, that if we know that there is exactly one change point or that there are no changes that are of equal size, then requiring this before updating our estimate for the change point is not necessary and can be omitted. This is written in Algorithm \ref{algo_step_1cp} lines 5-6.

\subsubsection{Stopping time}
We now present the second component of CPI, the stopping rule.
We give an explicit definition for the stopping time and show that a policy using it satisfies the first condition (equation \eqref{eqn_1cp_defn1}) of Definition \ref{defn_exact1_objective} for an Exact-$(1,\delta)$ policy. 
%%%%%%%%%%%%%%%%%%%%%%%%%%%%%%
% In previous works using Track-and-Stop methods,
% %\citep[such as][]{Degenne2019NonAsymptoticPE}
% a Chernoff stopping time was used \citep{kaufmann_16a}. This stopping time stops the algorithm when an infimum of a self-normalized sum \citep{kaufmann_martingales,BanditAlgosBook} crosses some threshold $\beta$. The general form of these stopping times are
% \begin{equation}\label{eqn_stopping_time_general_form}
%     \min\left\{t:\inf_{v' \in V^{alt}_{K,1}(\hat{x})} \sum_{i=1}^K T_i(t) D\left(v'(i),\hat{v}(i)\right) \geq \beta(t,\delta)\right\}
% \end{equation}
% where $\hat{v}$ is an estimate for the environment.
% However, in our setting since it is optimal to only play either side of the change point, we demonstrate that it is sufficient to only consider the self-normalized sum over actions $ i \in \{\hat{x}_t,\hat{x}_t+1\}$ since we expect the contribution from these terms to dominate. In Proposition \ref{prop_stopping_time} we explicitly define our proposed stopping time and show that it provides the desired level of confidence for our policy to be Exact-$(1,\delta)$. 
% %
%%%%%%%%%%%%%%%%%%%%%%%%%%%%%%%%%
We emphasize that, unlike in many prior track-and-stop works, the condition for the stopping time can also be checked explicitly and directly, rather than needing to solve the optimization problem as in \eqref{eqn_stopping_time_general_form1} \citep[e.g.][]{Yang2022OptimalClustering,kaufmann_16a}. In particular, for our proposed stopping time defined in Proposition \ref{prop_stopping_time}, we only need to check if the threshold has been exceeded in \eqref{eqn_defn_stopping_time}. No numerical optimization is required for checking this in each round, making it computationally efficient. This makes our proposed approach simple and further computationally efficient.

\begin{proposition}\label{prop_stopping_time}Define the stopping time in Algorithm \ref{algo_TandS} as
    \begin{equation} \label{eqn_defn_stopping_time}
    \tau_\delta =  \min\left\{ t:  \frac{T_{\hat{x}_t}(t)T_{\hat{x}_t+1}(t)}{2(T_{\hat{x}_t}(t)+T_{\hat{x}_t+1}(t))}      \hat{\Delta}^2_{\hat{x}_t} \geq \beta(t,\delta) \right\}
\end{equation}
with 
\begin{equation*}
    \hat{\Delta}^2_{\hat{x}_t} := |\hat{\mu}_{\hat{x}_t}(t) - \hat{\mu}_{\hat{x}_t+1}(t)|^2.
\end{equation*}
We define the threshold
\begin{equation}\label{eqn_beta_definition}
    \beta(t,\delta) = \log\left(t\gamma(K-1)/\delta\right)+ 8\log\log(t\gamma(K-1)/\delta)
\end{equation}
with $\gamma=2e^3 9^6/\log(3)$. Then in any environment $v \in V_{K,N}$ where $N \geq 1$ and any policy $\pi$ with stopping time $\tau_\delta$, we have
\begin{equation} 
    \mathbb{P}_{\pi,v}(\hat{x}_{\tau_\delta}\notin \underline{x}^*_v) \leq \delta.
\end{equation}

\end{proposition}
\begin{proof}
    % See APPENDIX - \textcolor{red}{proably helpful to describe what we discussed above in that we need only consider the two actions either side}
     For a full proof see Appendix \ref{app_proof_prop_stopping_time}. 
    This stopping time is motivated by Chernoff stopping times \citep{kaufmann_martingales} of the general form
    \begin{equation}\label{eqn_stopping_time_general_form1}
    \min\left\{t:\inf_{v' \in V^{alt}_{K,1}(\hat{x})} \sum_{i=1}^K T_i(t) D\left(v'(i),\hat{v}(i)\right) \geq \beta(t,\delta)\right\}.
\end{equation}
%We note that in our setting, s
Since it is optimal to only play either side of the change point, we demonstrate that it is sufficient to only consider the sum in \eqref{eqn_stopping_time_general_form1} over actions $ i \in \{\hat{x}_t,\hat{x}_t+1\}$.
\end{proof}
%Note that we will demonstrate later that the second part of our objectives also holds, namely the algorithm will stop almost surely using this stopping time.

%\textcolor{red}{While the stopping time displayed here excludes potential information from actions further away from the change point, we will show that this form is actually most appropriate for ANY objective described in the next section}

Interestingly, the form of the stopping time in \eqref{eqn_defn_stopping_time} is reminiscent of common tests in offline change point analysis literature \citep[e.g. see the GLR test statistic CUSUM seen in][]{ChenGuptaCP_Book, Verzelen2020OptimalChangeDetection_localization}. 
%We also point out that the form makes it seem amenable to an extension to multiple change points, this will be discussed in more detail in Section~\ref{section:multiple_cp}. 
%
%
Our algorithm CPI is given explictly in Algorithm \ref{algo_TandS}, where we denote 
\begin{equation}\label{eqn_Zt_definition}
    Z(t) =\frac{T_{\hat{x}_t}(t)T_{\hat{x}_t+1}(t)}{2(T_{\hat{x}_t}(t)+T_{\hat{x}_t+1}(t))}      \hat{\Delta}^2_{\hat{x}_t}.
\end{equation}

\begin{figure}[t]
\vspace{-1.5em}
\begin{algorithm}[H]
\caption{Single Change Point Identification (CPI)}\label{algo_TandS}
\begin{algorithmic}[1]
\STATE {\bfseries Input:} confidence $\delta \in (0,1)$ 
\STATE Play each action once
\WHILE{$Z(t) < \beta(t,\delta)$} %\COMMENT{Stopping time \eqref{eqn_defn_stopping_time}}
\STATE Update $\hat{\mu}_i(t)$, $\hat{\alpha}_t$, $\hat{x}_t$ 
\IF{$\text{min}_{i\in \{1,\dots, K\}} T_i(t) < \sqrt{t}$}
\STATE Play action $a_t = \text{argmin}_{i\in \{1,\dots, K\}} T_i(t)$ 
\ELSE
\STATE Play $a_t = \text{argmin}_{i \in \{\hat{x}_t,\hat{x}_t+1\}} T_i(t) $
\ENDIF
\ENDWHILE
\STATE \textbf{Return:} $\hat{x}_\tau$
\end{algorithmic}
\end{algorithm}
\vspace{-2.5em}
\end{figure}

\subsection{Asymptotic Upper Bound}
\def\mycmd{2} %can swap between two versions of a section with this...
\if\mycmd 1
\subsubsection{Non-asymptotic Upper Bound}
We now prove a non-asymptotic upper bound for the sample complexity of CPI (Algorithm \ref{algo_TandS}). The broad idea for the analysis is to firstly prove that: if the empirical mean reward from each action is well concentrated, then, due to the forced exploration, our estimates for the change point and optimal proportions ($\hat{x}_t,\hat{\alpha}_t$) should get closer to their true values ($x^*, \alpha^*$) over time. Hence, due to the tracking, the number of times we have played actions $x^*$ and $x^*+1$ adjacent to the change point should quickly increase and get closer to $t/2$. Subsequently, the values of $Z(t)$
in our stopping time should increase quickly, passing the threshold $\beta$ and stopping the algorithm. The analysis for our upper bounds focuses on understanding the rate at which this 
%sequence of effects 
occurs. 

Due to the explicit solutions found for the optimization problems \eqref{eqn_1cp_opt_problem} and \eqref{eqn_stopping_time_general_form}, we are able to construct simple and interpretable non-asymptotic upper bounds for the sample complexity of algorithm CPI.
Our upper bounds depend on the quantities, 
\begin{align}
    r(t) &= \sqrt{\frac{4\log(t) + 2\log(2\log(t)) + 1/2}{t}}\label{eqn_r_defn}\\
    T_0(\delta) &= \min\left\{T \in \mathbb{N}^+: T \geq \frac{8\beta(T,\delta)}{(\Delta_{1}-r(T))^2} \right\}\nonumber\\
    T_1(v) &= \min \left\{T \in \mathbb{N}^+ : r(T) < \frac{\Delta_{1}}{2}\right\}\nonumber
\end{align}
where the stopping threshold $\beta(T,\delta)$ is defined in \eqref{eqn_beta_definition}.
%If there is exactly one change point in the environment we define $\Delta_{(2)}=0$, and let $\ell  = \text{argmax}_{i > 1} \{\Delta_{(i)}: \Delta_{(1)} > \Delta_{(i)}\}$.  
Intuitively, $T_1(v)$ represents the time taken for forced exploration to provide a good estimate for the position of the change point. Additionally, $T_0(\delta)$ represents the time taken for our tracking to put enough samples around the true change point for us to be confident in its position and stop. Using these definitions we state a high probability upper bound for the sample complexity of Algorithm 
\ref{algo_TandS}.
See Appendix \ref{app_proof_prop_complexity_UB_HP} for a proof of Proposition \ref{prop_complexity_UB_HP}.

% We prove a high probability upper bound for the sample complexity in Proposition \ref{prop_complexity_UB_HP}. This allows us to prove an upper bound for the expected sample complexity in Proposition \ref{prop_complexity_UB}. The analysis is outlined in more detail in APPENDIX.

\begin{proposition}\label{prop_complexity_UB_HP} 
    For any environment $v\in V_{K,1}$ and any $T > \max\{T_0(\delta), T_1(v)\}$, the stopping time from Algorithm \ref{algo_TandS} 
    satisfies
    %is only larger than $T$ with probability less than the following.
    $$\mathbb{P}(\tau_\delta >T) \leq 2eK\frac{\log(T)}{T^2}$$
        % \leq BT\exp(-CT^{1/8})$$
        This ensures soundness of our stopping time/algorithm, in particular $\mathbb{P}(\tau_\delta < \infty)=1$.
\end{proposition}

The above proposition shows that Algorithm \ref{algo_TandS} will stop almost surely, hence CPI satisfies \eqref{eqn_1cp_defn2}. Therefore, since our stopping time ensures \eqref{eqn_1cp_defn2} (Prop \ref{prop_stopping_time}), CPI is an Exact-$(1,\delta)$ policy.
%Hence Algorithm \ref{algo_TandS} also satisfies the second part of our objectives. 
Furthermore, using the high probability upper bound on the sample complexity of CPI in Proposition \ref{prop_complexity_UB_HP}, we can construct an upper bound for the expected sample complexity of CPI as follows in Proposition \ref{prop_complexity_UB}.

\begin{proposition}\label{prop_complexity_UB} For any environment $v\in V_{K,1}$, the expected sample complexity for Algorithm \ref{algo_TandS} is bounded by
        $$\mathbb{E}[\tau_\delta] \leq T_0(\delta)+ T_1(v) + 2eK.$$
\end{proposition}

\subsubsection{Asymptotic Upper Bound} \label{subsection:1cp_asymp_ub}
From Proposition \ref{prop_complexity_UB}, we see that the only dependence on $\delta$ in the upper bound for the expected sample complexity comes from the $T_0(\delta)$ term. Hence, by showing that $T_0(\delta)/\log(1/\delta)$ approaches the lower bound as $\delta \rightarrow 0$, we can show that our algorithm is asymptotically optimal. This is done in the following Theorem, whose complete proof is in Appendix~\ref{app_proof_TAS_asymp_UB}

\begin{theorem} \label{TAS_asymp_UB} For any $v \in V_{K,N}$ where $N \geq 1$, using algorithm $\pi$ described in Algorithm \ref{algo_TandS}, we can upper bound the expected sample complexity as
    $$\limsup_{\delta \rightarrow 0} \frac{\mathbb{E}_{\pi,v}[\tau_\delta]}{\log(1/\delta)} \leq c^*(v).$$
\end{theorem}

This demonstrates that CPI (Algorithm \ref{algo_TandS}) is an asymptotically optimal Exact-$(1\delta)$ policy according to the lower bound in Corollary \ref{theorem_exact_1cp_lb_explicit}. We additionally emphasize that this optimality is tight in terms of the constants that appear in both the asymptotic upper bound and the lower bounds.
\fi

%%%%%%%%%%%%%%%%%%%%%%%%%%%%%ONE VERSION

\if\mycmd 2

We prove an asymptotic upper bound for the expected sample complexity of CPI (Algorithm \ref{algo_TandS}). The broad idea for the analysis is to firstly prove that: if the empirical mean reward from each action is well concentrated, then, due to the forced exploration, our estimates for the change point and optimal proportions ($\hat{x}_t,\hat{\alpha}_t$) should get closer to their true values ($x^*, \alpha^*$) over time. Hence, due to the tracking, the number of times we have played actions $x^*$ and $x^*+1$ adjacent to the change point should quickly increase and get closer to $t/2$. Subsequently, the values of $Z(t)$
in our stopping time should increase quickly, passing the threshold $\beta$ and stopping the algorithm. The analysis for our upper bounds focuses on understanding the rate at which this 
%sequence of effects 
occurs. We state the upper bound in Theorem \ref{TAS_asymp_UB} with the proof detailed in Appendix \ref{Appendix_proof_TAS_asymp_UB}.

\begin{theorem} \label{TAS_asymp_UB} For any $v \in V_{K,1}$, using algorithm $\pi$ described in Algorithm \ref{algo_TandS}, we can upper bound the expected sample complexity as
    $$\limsup_{\delta \rightarrow 0} \frac{\mathbb{E}_{\pi,v}[\tau_\delta]}{\log(1/\delta)} \leq c^*(v).$$
\end{theorem}

This demonstrates that CPI (Algorithm \ref{algo_TandS}) is an asymptotically optimal Exact-$(1,\delta)$ policy according to the lower bound in Corollary \ref{theorem_exact_1cp_lb_explicit}. We emphasize that this optimality is tight in terms of the constants that appear in both the asymptotic upper bound and the lower bounds. In Section \ref{section:multiple_cp} we will illustrate that, due to the explicit solutions found for the optimization problems \eqref{eqn_1cp_opt_problem} and \eqref{eqn_stopping_time_general_form1}, we are able to construct simple and interpretable non-asymptotic upper bounds for the sample complexity of our methods as well.

\fi

%%%%%%%%%%%%%%%%%%%%%%%%%%%%%%%%%%%%%%%%%%%%%%%%%%%%%%%%%%%%%%%%%%%%%%%%%%%%%%%
%%%%%%%%%%%%%%%%%%%%%%%%%%%%%%%%%%%%%%%%%%%%%%%%%%%%%%%%%%%%%%%%%%%%%%%%%%%%%%%
% MULTIPLE CHANGE POITNS
%%%%%%%%%%%%%%%%%%%%%%%%%%%%%%%%%%%%%%%%%%%%%%%%%%%%%%%%%%%%%%%%%%%%%%%%%%%%%%%
%%%%%%%%%%%%%%%%%%%%%%%%%%%%%%%%%%%%%%%%%%%%%%%%%%%%%%%%%%%%%%%%%%%%%%%%%%%%%%%
\section{Identifying Multiple Change Points} \label{section:multiple_cp}
While identifying a single change point is important, in practice we may need to confidently identify $N$ change points. Hence, we extend our results and methods from Section \ref{section:one_cp} to consider multiple change point identification.
\subsection{Known number of changes}\label{subsection_known_exact_N}
% \begin{itemize}
%     \item Explain the objective of identifying m change points given that we know that there are exactly m change point (maybe write in a definition and give motivations/applications)
%     \item Give lower bound which has the inf displayed
%     \item Give lower bound which has a lower/upper bound for the inf
%     \item explain why this lower bound is loose and why it is harder to find the exact solution of the inf. Explain why we suspect that we cannot quite recover the sum of the individual complexities...
% \end{itemize}

Suppose that we are in an environment with exactly $N$ change points and the learner is given this information. In this case, we want to return an estimate for the set of $N$ change points which is equal to the true set of change points with probability greater than $1-\delta$. We refer to a policy that does this as an Exact-$(N,\delta)$ policy.
\begin{definition} \label{defn:multiple_exact}
    \textbf{Exact-$(N,\delta)$ policy} 
    % For any $v \in V_{K,N}$, our policy $\pi$ should return an estimate for the set of change points $\hat{\underline{x}}_\tau$ of size $N$ satisfying 
    For any $v \in V_{K,N}$, an Exact-$(N,\delta)$ policy $\pi$ with stopping time $\tau$ returns an estimate $\hat{x}_\tau$ satisfying 
\begin{align*}
    \mathbb{P}_{\pi,v} (\hat{\underline{x}}_\tau = \underline{x}^*_v) &> 1-\delta,\\
    \mathbb{P}_{\pi,v} (\tau < \infty) &=1.
\end{align*}
\end{definition}

By starting with a general bound (as in Theorem \ref{theorem_1cp_general}), we can prove the following lower bound on the expected sample complexity of any Exact-$(N,\delta)$ policy.

% Furthermore, we can derive the following instance dependent lower bound for policies that satisfy this objective.

\begin{theorem}\label{theorem_exact_m_lb}
    For any Exact-$(N,\delta)$ policy $\pi$ with stopping time $\tau$, a lower bound for the expected stopping time in environment $v \in V_{K,N}$ is 
    \begin{align}
        \mathbb{E}_{\pi,v}[\tau] &\geq c^*_2(v) \log\left(\frac{1}{4\delta}\right) \label{eqn_theorem_exact_m_lb1}\\
        &\geq 4\sigma^2 \log\left(\frac{1}{4\delta}\right) \left(\sum_{i=1}^{N} \frac{1}{\Delta_i^2}\right), \label{eqn_theorem_exact_m_lb2}
    \end{align}
    where we define
    \begin{equation} \label{eqn_mcp_opt_problem}
        c^*_2(v)^{-1} =  \sup_{\alpha \in \mathcal{P}_{K}}\inf_{v' \in V^{alt}_{K,N}(\underline{x}^*_v)} \sum_{i=1}^K \alpha_i D(v_i,v_i').
    \end{equation}
    Here $V^{alt}_{K,N}(x^*_v)$ denotes the set of environments in $V_{K,N}$ whose set of change points are not equal to $\underline{x}^*_v$ and $\mathcal{P}_{K}$ is the standard $K$-dimensional simplex.
\end{theorem}

If the learner knows that there is exactly $N$ change points, then knowledge regarding the location of one change point can be informative for the location of an adjacent change point. 
Incorporating this information means that the optimal proportion of actions we should play near one change point no longer just depends on that change point, but also on adjacent change points. Because of this, the optimization problem in equation \eqref{eqn_mcp_opt_problem} becomes more complicated and finding a general closed form solution becomes more challenging than the Exact-$(1,\delta)$ case in \eqref{eqn_1cp_opt_problem} where we do not have this coupled effect. 
%However, Theorem 5.2 tells us that the cost in sample complexity of not knowing the true number of change points is at most a factor of 2 asymptotically. 
%Moreover, even when the number of change points is known, we can run the same computationally efficient algorithm, MCPI. This algorithm avoids solving the complex optimization problem and obtains sample complexity within a factor of 2 of the optimal (non-closed form) rate asymptotically. 
%%%% Hence, solving the optimization problem \eqref{eqn_mcp_opt_problem} explicitly is much more challenging than the Exact-$(1,\delta)$ case in \eqref{eqn_1cp_opt_problem} where we do not have this coupled effect. 
%In Theorem \ref{theorem_exact_m_lb} the optimization problem of finding $c^*_2(v)$ is significantly more challenging than in Theorem \ref{theorem_exact_1cp_lb_explicit}, hence we are unable to solve it explicitly as before. 
However, we are able to lower bound $c^*_2(v)$ to provide the final lower bound \eqref{eqn_theorem_exact_m_lb2} in Theorem \ref{theorem_exact_m_lb}. 
Note that the lower bound in \eqref{eqn_theorem_exact_m_lb2} is the sum of the complexities of finding a single change point (shown in Theorem \ref{theorem_exact_1cp_lb_explicit}) in $N$ different settings, up to a factor of 2. 
% Furthermore, as we will see from the rest of Section \ref{section:multiple_cp}, Theorem \ref{theorem_exact_m_lb} tells us that the cost in sample complexity of not knowing the true number of change points is at most a factor of 2 asymptotically.
Furthermore as we will see from the rest of Section \ref{section:multiple_cp}, compared to settings where we do know the true number of change points (Theorem \ref{theorem_exact_m_lb}), the cost of not knowing the true number of change points is asymptotically at most a factor of two in the expected stopping time (Theorems \ref{theorem_any_m_lb_specific_case}, \ref{theorem_any_m_lb_general_case}, and \ref{theorem_upper_mcp_asymptotic}).
%We emphasize that in the case where we know the number of change points, the complexity of the problem is not simply the sum of the complexities of identifying the individual change points. 
%This is because data demonstrating the position of some change points can be informative on the position of other change points.
% 
% While we could directly use a track-and-stop style approach here by approximating the optimal proportions by numerically solving optimization problems of the form \ref{eqn_mcp_opt_problem} in each round, this would be computationally very intensive. Whereas later we will provide a computationally efficient method which is optimal up to a constant 2.

\subsection{Unknown number of changes}\label{subsection_unknown_num_changes_mcp}
% \begin{itemize}
%     \item Explain objective of identifying m change points without knowledge of how many change points there are (only the knowledge that there is at least m changes)... 
%     \item Give lower bound for unknown number of change points in an environment where there happens to be m change points (mentioning the optimal proportions from this and how it compares to the one change point setting)
%     \item Give lower bound for unknown number of change points in an environment where there is at least m change points and pointing out how this compares to the previous lower bound and intuition.
%     \item Comment on the similarity with lower bound above, particularly as delta goes to zero. Explain intuition. Explain that we can get an algorithm which is also asymptotically optimal for this objective and only suboptimal by at most 1/2 for the above objective simultaneously. (maybe instead of suboptimal say optimal up to 1/2...)
% \end{itemize}

When we want to confidently identify $N$ change points in our environment, it is important to provide methods which are robust to the presence of additional change points. In particular, we would like to develop policies which are able to identify $N$ change points out of an \emph{unknown number of change points}. In this case, we want to return an estimate for a set of $N$ change points which is in the set of true change points $\underline{x}^*_v$, where $|\underline{x}^*_v| = m \geq N$, with probability greater than $1-\delta$. We refer to a policy that achieves this as an Any-$(N,\delta)$ policy.
% We can also define the objective of finding any $m$ change points within the set of true change points with high probability as follows.

\begin{definition} \label{defn:multiple_any}
    \textbf{Any-$(N,\delta)$ Policy} 
    For any $v \in V_{K,m}$ where $m \geq N$, 
    %our policy $\pi$ should return an estimate for the set of change points $\hat{\underline{x}}_\tau$ of size $m$ satisfying 
    %For any $v \in V_{K,m}$, 
    an Any-$(N,\delta)$ policy $\pi$ with stopping time $\tau$ returns an estimate $\hat{\underline{x}}_\tau$ of size $N$ satisfying
\begin{align}
    \mathbb{P}_{\pi,v} (\hat{\underline{x}}_\tau \subseteq \underline{x}^*_v) &> 1-\delta,\label{defn_any1_objective}\\
    \mathbb{P}_{\pi,v} (\tau < \infty) &=1.\label{defn_any2_objective}
\end{align}
\end{definition}
Suppose that we had an Any-$(N,\delta)$ policy $\pi$ and, for now, additionally suppose that there are exactly $N$ change points in the environment (note that this is different to the Exact-$(N,\delta)$ setting considered in Section~\ref{subsection_known_exact_N} where the learner knows there are exactly $N$ change points). 
%Now, suppose that we were to try and find any $N$ change points in an environment where there happens to be exactly $N$ change points in our environment. 
In this case, we can show that the optimal proportions become 
% \begin{equation}\label{eqn_M_geq_m_alpha_star}
%     \alpha^*_{x^*_{v,j}} = \alpha^*_{x^*_{v,j}+1}=\frac{\frac{1}{\Delta_j^2}}{2\sum_{i=1}^{m} \frac{1}{\Delta_i^2}}
% \end{equation}
\begin{equation}\label{eqn_M_geq_m_alpha_star}
    \alpha^*_{j} = \alpha^*_{j+1}=\frac{\frac{1}{\Delta_j^2}}{2\sum_{i=1}^{N} \frac{1}{\Delta_i^2}}
\end{equation}
for $j \in \{x^*_{v,1},\dots,x^*_{v,N}\}$, and zero for all other $\alpha^*_j$ values. 
This intuitively suggests that constructing an optimal policy requires (asymptotically) sampling most of our actions adjacent to the changes in mean. The proportion of samples around each change in mean should be inversely proportional to the size of the change squared.
This allows us to prove the lower bound in Theorem \ref{theorem_any_m_lb_specific_case} below, detailed in Appendix \ref{app_proof_theorem_any_m_lb_specific_case}. 

\begin{theorem}\label{theorem_any_m_lb_specific_case}
    For any Any-$(N,\delta)$ policy $\pi$ with stopping time $\tau$, a lower bound for the expected stopping time in environment $v \in V_{K,N}$ is
    \begin{equation*}
        \mathbb{E}_{\pi,v}[\tau] \geq 8\sigma^2 \log\left(\frac{1}{4\delta}\right) \left(\sum_{i=1}^{N} \frac{1}{\Delta_i^2}\right).
    \end{equation*}
\end{theorem}

Now, suppose our aim is to confidently identify $N$ change points out of a set of $m \geq N$ change points. In this case, intuitively the complexity of finding any $N$ change points is lower bounded by the complexity of finding the $N$ most easily identifiable change points, $\{x^*_{(1)}, \dots, x^*_{(N)}\}$. This is formalized in Theorem \ref{theorem_any_m_lb_general_case} where the first sum in \eqref{eqn_theorem_any_lb_general} is over the $N$ largest changes in mean. 
% In this case, we can intuitively extend \eqref{eqn_M_geq_m_alpha_star} suggesting that we should optimally play most of our actions adjacent to the $N$ largest changes in mean. 
% See Appendix \ref{app_proof_theorem_any_m_lb_general_case} for full proof details.
See Appendix~\ref{app_proof_theorem_any_m_lb_general_case} for a proof.
%Now, more generally, if we are trying to identify any $m$ change points in the set of true change points. Then in order to provide an efficient algorithm, intuitively we should allocate our sampling efforts around the $m$ most significant change points which are easiest to identify. This can be formalized as follows.

\begin{theorem}\label{theorem_any_m_lb_general_case}
    For any Any-$(N,\delta)$ policy $\pi$ with stopping time $\tau$, the expected stopping time 
    % in environment $v \in V_{K,N}$ is
    % Suppose that a policy $\pi$ with stopping time $\tau$ satisfies the Any-$(m,\delta)$ Objective. Then a lower bound for the expected stopping time
    in environment $v \in V_{K,m}$ for $N \leq m <K$ is lower bounded by
    \begin{align}
        \mathbb{E}_{\pi,v}[\tau] \geq &8\sigma^2 (1-\delta)\log\left(\frac{1}{4\delta}\right) \left(\sum_{i=1}^{N} \frac{1}{\Delta_{(i)}^2}\right)\nonumber\\
        & - \log (2) \left(\sum_{i=1}^{m} \frac{1}{\Delta_i^2}\right).\label{eqn_theorem_any_lb_general}
    \end{align}
\end{theorem}

Note that, as $\delta \rightarrow 0$ the lower bound in Theorem~\ref{theorem_any_m_lb_general_case} becomes similar to the lower bound for the Exact-$(N,\delta)$ policies in Theorem \ref{theorem_exact_m_lb} up to a constant factor of 2. Furthermore, the lower bound in Theorem~\ref{theorem_any_m_lb_general_case} becomes similar to Theorem \ref{theorem_any_m_lb_specific_case}, except we are summing over the $N$ largest change points 
%rather than all the change points. 
rather than all of them.
In Section~\ref{subsection_mcp_algo_ubs} we provide a policy
which is simultaneously Exact-$(N,\delta)$ and Any-$(N,\delta)$. This policy is asymptotically optimal when the true number of change points is unknown ($m\geq N$) and optimal up to a factor of $2$ when the learner is given the exact number of true change points in the environment.
% to satisfy both the Exact-$(m,\delta)$ Objective and Any-$(m,\delta)$ Objective simultaneously. We will show that this same policy is asymptotically optimal for the Any-$(m,\delta)$ Objective and optimal up to a factor of 2 for the Exact-$(m,\delta)$ Objective. 

\subsection{Sequential Approach and Upper Bounds}\label{subsection_mcp_algo_ubs}
% \begin{itemize}
%     \item Explain that we could do a normal track and stop approach, mention this is in appendix, but mention it could be better to find one change point at a time (since similar to CP methods, since easier to extend to epsilon big objective, since they are both optimal, since if there is one change point significantly larger than all others for example normal TAS would find all cps at the same rate, whereas the proposed method would promote finding the most significant/easiest change points fastest.. a practitioner may find this useful.), etc.
%     \item Go through three main things of sampling rule, stopping rule, and recommendation rule - giving intuition
%     \item Mention algorithm will satisfy both objectives
%     \item Formally write algorithm here
%     \item Theorem: non-asymptotic upper bound
%     \item Theorem: asymptotic upper bound and comparison with lower bounds
% \end{itemize}

Motivated by Binary Segmentation methods from offline change point analysis \citep[e.g.][]{Fryzlewicz_2014_wild_binary_segmentation,original_binary_segmentation_paper} and and the single change point setting studied in Section \ref{section:one_cp}, we propose the Multiple Change Point Identification algorithm (MCPI, Algorithm \ref{algo_TandS2}). MCPI works by sequentially identifying one change point at a time until we have found $N$. In particular we repeatedly run the loop in CPI (Algorithm \ref{algo_TandS} lines 3-10), which identifies one change point, a total of $N$ times. After each loop, we will have stopped and identified one change point. We then remove this action from the set of potential change points and add this action to the set of change points we will return at the end (Algorithm \ref{algo_TandS2} line 17). After this we repeat the process until we have identified $N$ change points in total.

\textbf{Edge Case (Multiple Equally Sized Changes)}\quad Suppose we had an Any-$(1,\delta)$ policy, which confidently identifies one change point, and there are multiple equally sized changes in the environment. For example $\Delta_{v,(1)}=\Delta_{v,(2)}$. We see from Theorem \ref{theorem_any_m_lb_general_case} that the lower bound only includes the complexity from finding the position of any single \emph{one} of the largest changes. Hence, an optimal policy would simply focus its sampling efforts around any \emph{one} of the largest changes. By simply running CPI in this edge case, there is a risk that our estimate $\hat{x}_t$ will fluctuate between different change points of equal size and therefore our actions will not be focused around just one change point. To avoid this, in MCPI we include a condition that we only update our estimate $\hat{x}_t$ when one empirical change in mean reward is sufficiently larger than another. In particular, when there exists $i , j \in [K-1]$ such that 
\begin{equation}\label{eqn_edge_case_condition}
    |\hat{\mu}_i(t) - \hat{\mu}_{i+1}(t)| > |\hat{\mu}_j(t) - \hat{\mu}_{j+1}(t)| + r(t)
\end{equation}
holds (with $r(t)$ defined in \eqref{eqn_r_defn}), we update our estimated change point to be $\hat{x}_t (S)$. If \eqref{eqn_edge_case_condition} does not hold in round $t$ then we do not update our estimate, namely $\hat{x}_{t} \leftarrow \hat{x}_{t-1}$.
This is included in  lines 8-9 of MCPI (Algorithm \ref{algo_TandS2}).
Note however, that if we know that there is exactly one change point or that there are no changes of equal size, then we do not need to check \eqref{eqn_edge_case_condition} before updating our estimate for the change point and this can be omitted. 

% Note that an alternative approach to identifying $N$ change points could be to extend CPI by directly tracking (see \eqref{eqn_tracking}) with respect to our optimal proportions \eqref{eqn_M_geq_m_alpha_star}.
% %, similar to \eqref{eqn_tracking}. 
% However, this will lead to identifying all of the change points at the same rate. Whereas if there is one change point that is significantly larger than the rest, it may be advantageous for a practitioner to quickly identify this change first. This is what the proposed MCPI algorithm does.

% While it may be optimal to directly propose a track-and-stop procedure to directly track and asymptotically match the optimal proportions shown in \eqref{eqn_M_geq_m_alpha_star}, there are other optimal methods which may be more appropriate. The problem with matching these optimal proportions using direct tracking, is that this will lead to identifying all of the change points at the same rate. However, if there is one significantly larger change point than the rest, then it may be advantageous for a practitioner to quickly identify this change first. Hence, we can suggest an alternative method by repeatedly searching for one change point at a time. In particular, we propose running Algorithm \ref{algo_step_1cp} a total of $m$ times. After each run, we have stopped and identified one change point. We should then remove this action from the set of potential change points and continue. We will refer to this as MCPI and is written explicitly in Algorithm \ref{algo_TandS2}.

\begin{figure}[t]
\vspace{-1.5em}
\begin{algorithm}[H]
\caption{Multiple Change Point Identification (MCPI)}\label{algo_TandS2}
\begin{algorithmic}[1]
\STATE {\bfseries Input:} confidence $\delta \in (0,1)$
\STATE {\bfseries Input:} number of changes to look for, $N$
\STATE Initialize $S \leftarrow [K-1]$, $\underline{\hat{x}}_\tau \leftarrow \emptyset$
\STATE Play each action once
\FOR{phase $j$ in $\{1, \dots, N\}$}
\WHILE{$Z(t) < \beta(t,\delta/N)$} %\Comment{Stopping time \eqref{eqn_defn_stopping_time}}
\STATE Update $\hat{\mu}_i(t)$, $\hat{\alpha}_t$
\IF{$\exists i, j \in [K-1]$ s.t. \eqref{eqn_edge_case_condition} holds}
\STATE Update $\hat{x}_t =  \hat{x}_t(S)$ %\Comment{Simple estimator \eqref{eqn_estimator_simple_defn}}
\ENDIF
\IF{$\text{min}_{i\in \{1,\dots, K\}} T_i(t) < \sqrt{t}$}
\STATE Play action $a_t = \text{argmin}_{i\in \{1,\dots, K\}} T_i(t)$ 
\ELSE
\STATE Play $a_t = \text{argmin}_{i \in \{\hat{x}_t,\hat{x}_t+1\}} T_i(t) $
\ENDIF
\ENDWHILE
\STATE $\underline{\hat{x}}_\tau \leftarrow \underline{\hat{x}}_\tau\cup \hat{x}_t$ , $S \leftarrow S\backslash \{\hat{x}_t\}$ 
\ENDFOR
\STATE \textbf{Return:} $\underline{\hat{x}}_\tau$
\end{algorithmic}
\end{algorithm}
\vspace{-2em}
\end{figure}

% \begin{algorithm}[t]
% \caption{Multiple Change Point Identification (MCPI)}\label{algo_TandS2}
% \begin{algorithmic}[1]
% \State {\bfseries Input:} confidence $\delta \in (0,1)$
% \State {\bfseries Input:} number of changes to look for $N$
% \State Initialize $S \leftarrow [K-1]$, $\underline{\hat{x}}_\tau \leftarrow \emptyset$
% \State Play each action once
% \For{phase $j$ in $\{1, \dots, N\}$}
% \While{$Z(t) < \beta(t,\delta/N)$} \Comment{Stopping time \eqref{eqn_defn_stopping_time}}
% \State Update $\hat{\mu}_i(t)$, $\hat{\alpha}_t$
% \If{$\exists i, j \in [K-1]$ s.t. \eqref{eqn_edge_case_condition} holds}
% \State Update $\hat{x}_t(S)$ \Comment{Simple estimator \eqref{eqn_estimator_simple_defn}}
% \EndIf
% \If{$\text{min}_{i\in \{1,\dots, K\}} T_i(t) < \sqrt{t}$}
% \State Play action $a_t = \text{argmin}_{i\in \{1,\dots, K\}} T_i(t)$ 
% \Else
% \State Play $a_t = \text{argmin}_{i \in \{\hat{x}_t,\hat{x}_t+1\}} T_i(t) $
% \EndIf
% \EndWhile
% \State $\underline{\hat{x}}_\tau \leftarrow \underline{\hat{x}}_\tau\cup \hat{x}_t$ , $S \leftarrow S\backslash \{\hat{x}_{\tau,j}\}$ 
% \EndFor
% \State \textbf{Return:} $\underline{\hat{x}}_\tau$
% \end{algorithmic}
% \end{algorithm}

% Similar to the single change point setting in Section~\ref{section:one_cp}, we 
We now
provide a non-asymptotic upper bound for the expected stopping time of MCPI. To do so, we define  
% $$T_0'(\delta) = \min\left\{T \in \mathbb{N}^+: T -2KT^{\frac{1}{2}}\geq \sum_{i=1}^N\frac{8\beta(T,\delta/N)}{(\Delta_{(i)}-r(T))^2} \right\}$$
% $$T_1'(v) = \min \left\{T \in \mathbb{N}^+ : r(T) < \frac{\Delta_{(N)}- \Delta_{(\ell)}}{2}\right\}.$$

% \begin{align}
% r(t) &= \sqrt{\frac{4\log(t) + 2\log(2\log(t)) + 1/2}{(t^{1/4}-K)_+}}\label{eqn_r_defn}\\
%     T_0'(\delta) &= \min\left\{T \in \mathbb{N}^+: T -2KT^{\frac{1}{2}}\geq \sum_{i=1}^N\frac{8\beta(T,\delta/N)}{(\Delta_{(i)}-2r(T))^2} \right\}\nonumber\\
% T_1'(v) &= \min \left\{T \in \mathbb{N}^+ : r(T) < \frac{\Delta_{(N)}- \Delta_{(\ell)}}{4}\right\}.\nonumber
% \end{align}
\begin{equation}
    r(t) = \sqrt{\frac{4\log(t) + 2\log(2\log(t)) + 1/2}{(t^{1/4}-K)_+}}\label{eqn_r_defn}
\end{equation}
\begin{align}
    T_0'(\delta) &= \min\left\{T \in \mathbb{N}: T -2KT^{\frac{1}{2}}\geq \sum_{i=1}^N\frac{8\beta(T,\delta/N)}{(\Delta_{(i)}-2r(T))^2} \right\}\nonumber\\
T_1'(v) &= \min \left\{T \in \mathbb{N} : r(T) < \frac{\Delta_{(N)}- \Delta_{(\ell)}}{4}\right\}.\nonumber
\end{align}

Here $\beta$ is the threshold function defined in equation \eqref{eqn_beta_definition}.
%and $r(T)$ is defined in equation \eqref{eqn_r_defn}. 
If there are exactly $m$ change points in the environment we define $\Delta_{(m+1)}=0$. We also let $\ell  = \text{argmax}_{i > N} \{\Delta_{(i)}: \Delta_{(N)} > \Delta_{(i)}\}$ be such that $\Delta_{(\ell)}$ is the next largest change in mean \emph{strictly} smaller than $\Delta_{(N)}$. Intuitively, $T_1'(v)$ represents the time taken for forced exploration to provide a good estimate for the positions of the change points in environment $v$. Additionally, $T_0'(\delta)$ represents the time taken for our tracking to put enough samples around the true change point for us to be confident in its position and stop. Using these definitions 
we provide the following non-asymptotic upper bound for the expected stopping time of MCPI.

\begin{proposition} \label{prop_MCPI_ub_exp}
For any environment $v\in V_{K,m}$ for $ m \geq N$, the expected sample complexity for Algorithm \ref{algo_TandS2} is bounded by
        $$\mathbb{E}[\tau_\delta] \leq T_0'(\delta)+ T_1'(v) + 2eK.$$
\end{proposition}

Subsequently, by considering $\delta \rightarrow 0$, we provide an asymptotic upper bound for the sample complexity of MCPI and show that it is simultaneously Exact-$(N,\delta)$ and Any-$(N,\delta)$. %in Theorem \ref{theorem_upper_mcp_asymptotic} below. 
% We see that the sample complexity for MCPI asymptotically matches the lower bound for any Any-$(N,\delta)$ policy (Theorem \ref{theorem_any_m_lb_general_case}) with the correct constants. Furthermore we see that MCPI simultaneously matches the lower bound for any Any-$(N,\delta)$ policy (Theorem \ref{theorem_exact_m_lb}) asymptotically up to (at most) a constant factor of $2$. In the special case where there is exactly $N=1$ change point and this is known to the learner, MCPI is also optimal with correct constants (Theorem \ref{theorem_exact_1cp_lb_explicit}).

%We see that this is asymptotically optimal for the Any-$(m,\delta)$ Objective as the upper bound shown in Theorem \ref{theorem_upper_mcp_asymptotic} is equal to the lower bound shown in Theorem \ref{theorem_any_m_lb_general_case} as $\delta \rightarrow 0$ with the correct constants. Additionally, by comparing Theorem \ref{theorem_upper_mcp_asymptotic} with Theorem \ref{theorem_exact_m_lb} we see that MCPI is also asymptotically optimal for the Exact-$(m,\delta)$ objective up to (at most) a constant factor of $2$.

\begin{theorem} \label{theorem_upper_mcp_asymptotic}
MCPI (Algorithm \ref{algo_TandS2}) is an Exact-$(N,\delta)$ policy and an Any-$(N,\delta)$ policy. For any $v \in V_{K,m}$ where $m \geq N$, we can upper bound the expected sample complexity of MCPI as
    $$\limsup_{\delta \rightarrow 0} \frac{\mathbb{E}_{\pi,v}[\tau_\delta]}{\log(1/\delta)} \leq 8\sigma^2 \left(\sum_{i=1}^{N} \frac{1}{\Delta_{(i)}^2}\right).$$
\end{theorem}
We see that the sample complexity for MCPI asymptotically matches the lower bound for any Any-$(N,\delta)$ policy (Theorem \ref{theorem_any_m_lb_general_case}) with the correct constants. Furthermore we see that MCPI simultaneously matches the lower bound for any Exact-$(N,\delta)$ policy (Theorem \ref{theorem_exact_m_lb}) asymptotically up to (at most) a constant factor of $2$. 
Note that we could remove this constant by directly tracking and numerically approximating the optimal proportions from \eqref{eqn_mcp_opt_problem}, however this could be computationally very expensive and would not be robust to an unknown number of changes, unlike MCPI.
In the special case where there is exactly $N=1$ change point and this is known to the learner, MCPI is also optimal with correct constants (Theorem \ref{theorem_exact_1cp_lb_explicit}).

We note that another alternative approach to identifying any $N$ change points could be to extend CPI by directly tracking (as in \eqref{eqn_tracking}) with respect to an estimate of our optimal proportions \eqref{eqn_M_geq_m_alpha_star}.
%, similar to \eqref{eqn_tracking}. 
We suspect this alternative approach will also attain an optimal upper bound asymptotically.
However, this will lead to identifying all of the change points at the same rate. Whereas if there is one change point that is significantly larger than the rest, it may be advantageous for a practitioner to quickly identify this change first. This is what the proposed MCPI algorithm does.

%%%%%%%%%%%%%%%%%%%%%%%%%%%%%%%%%%%%%%%%%%%%%%%%%%%%%%%%%%%%%%%%%%%%%%%%%%%%%%%
%%%%%%%%%%%%%%%%%%%%%%%%%%%%%%%%%%%%%%%%%%%%%%%%%%%%%%%%%%%%%%%%%%%%%%%%%%%%%%%
% EXPERIMENTS
%%%%%%%%%%%%%%%%%%%%%%%%%%%%%%%%%%%%%%%%%%%%%%%%%%%%%%%%%%%%%%%%%%%%%%%%%%%%%%%
%%%%%%%%%%%%%%%%%%%%%%%%%%%%%%%%%%%%%%%%%%%%%%%%%%%%%%%%%%%%%%%%%%%%%%%%%%%%%%%
\section{Experiments} \label{section:experiments}
% \begin{itemize}
%     \item Comparison with proposed algorithm upper bound and lower bound \textcolor{blue}{(done)}
%     \item Comparison with proposed algorithm with clustering \textcolor{blue}{(done)}
%     \item Comparison between proposed methods and potential improvement when running the exact optimization
%     \item Maybe also simulate setting in which there are many more means and MCPI still matches the lower bound
% \end{itemize}

\begin{figure}[t]
\vskip -0.0in
\begin{center}
\centerline{\includegraphics[width=\columnwidth]{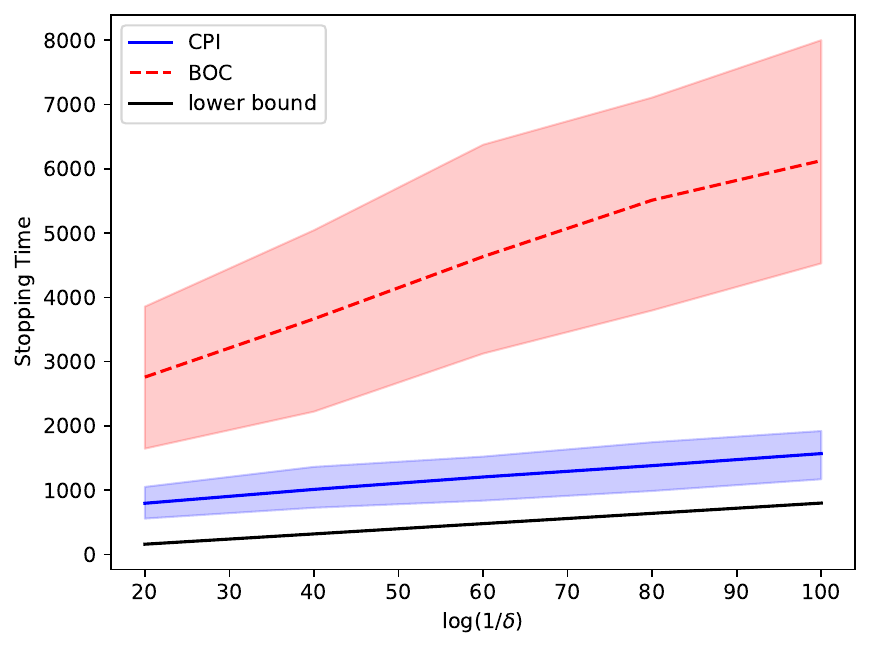}}
\caption{We run the MCPI and BOC algorithm at a range of values for $\delta$ in environment $v_1$. At each confidence level we repeat 100 runs and plot the average stopping time with 90 percent confidence intervals. We also plot the lower bound in Theorem \ref{theorem_any_m_lb_general_case}.}
\label{fig1}
\end{center}
\vskip -0.4in
\end{figure}

To complement our theoretical results we conduct experiments to test our proposed algorithms in synthetic environments. In Figure \ref{fig1}, we consider an environment $v_1$ with only one change point with means $\mu = (2,2,2,2,2,2,1,1,1)$.
% \begin{equation*}
%     \mu = \{2.0,2.0,2.0,2.0,2.0,2.0,1.0,1.0,1.0\}
% \end{equation*}
 In this case, we run our proposed policy MCPI (setting $N=1$) and a range of choices for $\delta$. At each of these choices for $\delta$ we run MCPI 100 times and plot the average stopping time. We also plot our lower bound from Theorem \ref{theorem_any_m_lb_general_case} on the same axes. From Figure \ref{fig1} we see that the our expected stopping time increases parallel to the lower bound, supporting our theoretical demonstration of the asymptotic optimality of MCPI. On the same figure we plot the stopping times when, instead of tracking with respect to our estimated optimal proportions \eqref{eqn_1cp_alpha_star}, we sample with respect to the oracle optimal proportions of the Bandit Online Clustering (BOC) algorithm \citep{Yang2022OptimalClustering}. In particular, we provide BOC with the true number of change points (required as an input of BOC) as well as the oracle optimal proportions $\alpha^*$ described in \citep{Yang2022OptimalClustering} for clustering in $v_1$. Despite the additional information provided to BOC, it does not take advantage of the piecewise constant structure of $v_1$. Hence we see that BOC's sample complexity increases at a much faster rate than MCPI and the lower bound, emphasizing the sub optimality of BOC in the fixed confidence piecewise constant bandits setting. See Appendix~\ref{app_sims} for further simulations with multiple change points.

%%%%%%%%%%%%%%%%%%%%%%%%%%%%%%%%%%%%%%%%%%%%%%%%%%%%%%%%%%%%%%%%%%%%%%%%%%%%%%%
%%%%%%%%%%%%%%%%%%%%%%%%%%%%%%%%%%%%%%%%%%%%%%%%%%%%%%%%%%%%%%%%%%%%%%%%%%%%%%%
% CONCLUSION
%%%%%%%%%%%%%%%%%%%%%%%%%%%%%%%%%%%%%%%%%%%%%%%%%%%%%%%%%%%%%%%%%%%%%%%%%%%%%%%
%%%%%%%%%%%%%%%%%%%%%%%%%%%%%%%%%%%%%%%%%%%%%%%%%%%%%%%%%%%%%%%%%%%%%%%%%%%%%%%
\section{Discussion and future work} \label{section:discussion}

% \begin{itemize}
%     \item Put in place a short summary of what we have presented and how this has contributed to the literature
%     \item Comment that we would like to extend this to a setting with a fully unknown number of change points and that we suspect this is possible using the existing method... i.e. if there was no clear intuition for how many change points to search for, then we could simply start the same procedure looking for one change point at a time, however we could additionally include th
% \end{itemize}

In this paper we have studied the fixed confidence piecewise constant bandits problem.
%the change point identification problem under bandit feedback in the fixed confidence setting. 
We proved multiple instance-dependent lower bounds on the expected stopping time of policies with different objectives, illustrating how the magnitude of the change points 
%and the noise variance 
affect the complexity of our problem. Additionally, we constructed the computationally efficient MCPI algorithm to sequentially locate $N$ change points with fixed confidence. By proving non-asymptotic and asymptotic upper bounds we showed that MCPI is asymptotically optimal under different objectives.
%We presented multiple lower bounds for the expected stopping time of policies under different objectives. We provided the intuitive methods CPI and MCPI for single and multiple change point settings respectively. Furthermore, we have provided non-asymptotic as well as asymptotic upper bounds for these policies which are asymptotically optimal under different objectives.

A related problem is the task of identifying \emph{all} changes greater than some $\epsilon>0$. This would be relevant in settings where there is no domain-specific or contextual knowledge of an appropriate number of change points to search for. We expect that it would be possible to extend MCPI to this setting, but we leave this to future work.
%For future work we suspect that it would be possible to extend MCPI in order to identify \emph{all} changes greater than some $\epsilon$ in settings where there is no domain-specific or contextual knowledge for an appropriate number of change points to search for. However this is beyond the scope of this paper.
Another direction for future work is to extend our setting to try to identify change points between piecewise smooth regions of a continuous action space (e.g. $\mathcal{A}=[0,1]$).
We could also consider different objectives such as minimizing the simple regret (i.e. the bias of our change point estimates) or maximizing the cumulative rewards in this piecewise constant setting.
For now, we have filled a gap in the change point identification literature by providing an in-depth study of the fixed-confidence multiple change point identification problem. 

% \def\mycmdd{1} %can swap between two versions of a section with this...
% \if\mycmdd 1
% \textcolor{red}{ways to save space: remove the discussion of self-normalized sums (1/4 page), potentially remove the discussion. Can also maybe get rid of the non-asymptotic stuff for CPI... (1/2 page) }
% \fi

%\newpage
%%%%%%%%%%%%%%%%%%%%%%%%%%%%%%%%%%%%%%%%%%%%%%%%%%%%%%%%%%%%%%%%%%%%%%%%%%%%%%%
%%%%%%%%%%%%%%%%%%%%%%%%%%%%%%%%%%%%%%%%%%%%%%%%%%%%%%%%%%%%%%%%%%%%%%%%%%%%%%%
% ACKNOWLEDGEMENTS
%%%%%%%%%%%%%%%%%%%%%%%%%%%%%%%%%%%%%%%%%%%%%%%%%%%%%%%%%%%%%%%%%%%%%%%%%%%%%%%
%%%%%%%%%%%%%%%%%%%%%%%%%%%%%%%%%%%%%%%%%%%%%%%%%%%%%%%%%%%%%%%%%%%%%%%%%%%%%%%
\section*{Acknowledgments}
The authors would like to thank Professor Niall Adams for support and insightful discussions about this problem. The authors would also like to thank the anonymous reviewers and area chairs for their feedback and suggestions during the review process.
Joseph Lazzaro was supported by a Roth Scholarship from the Department of Mathematics, Imperial College London.

\section*{Impact Statement}

This paper presents work whose goal is to advance the field of 
Machine Learning. There are many potential societal consequences 
of our work, none which we feel must be specifically highlighted here.

\newpage
\bibliography{bibliography}

\begin{thebibliography}{43}
\providecommand{\natexlab}[1]{#1}
\providecommand{\url}[1]{\texttt{#1}}
\expandafter\ifx\csname urlstyle\endcsname\relax
  \providecommand{\doi}[1]{doi: #1}\else
  \providecommand{\doi}{doi: \begingroup \urlstyle{rm}\Url}\fi

\bibitem[Aminikhanghahi \& Cook(2017)Aminikhanghahi and Cook]{CP_survey}
Aminikhanghahi, S. and Cook, D.
\newblock A survey of methods for time series change point detection.
\newblock \emph{Knowledge and Information Systems}, 2017.

\bibitem[Auer et~al.(2019)Auer, Gajane, and Ortner]{non_stationary_CP_auer}
Auer, P., Gajane, P., and Ortner, R.
\newblock Adaptively tracking the best bandit arm with an unknown number of distribution changes.
\newblock In \emph{Proceedings of the Thirty-Second Conference on Learning Theory}, 2019.

\bibitem[Bacchiocchi et~al.(2025)Bacchiocchi, Castiglioni, Marchesi, and Gatti]{bacchiocchi2025regret}
Bacchiocchi, F., Castiglioni, M., Marchesi, A., and Gatti, N.
\newblock Regret minimization for piecewise linear rewards: Contracts, auctions, and beyond.
\newblock In \emph{Proceedings of the 26th ACM Conference on Economics and Computation}, 2025.

\bibitem[Barrier et~al.(2022)Barrier, Garivier, and Koc\'ak]{GaussianBAI}
Barrier, A., Garivier, A., and Koc\'ak, T.
\newblock A non-asymptotic approach to best-arm identification for gaussian bandits.
\newblock In \emph{Proceedings of The 25th International Conference on Artificial Intelligence and Statistics}, 2022.

\bibitem[Bauer(1958)]{Bauer1958MinimalstellenVF_maximum_principle}
Bauer, H.
\newblock Minimalstellen von funktionen und extremalpunkte.
\newblock \emph{Archiv der Mathematik}, 9:\penalty0 389--393, 1958.

\bibitem[Bubeck et~al.(2011)Bubeck, Munos, Stoltz, and Szepesvári]{treeAlgo}
Bubeck, S., Munos, R., Stoltz, G., and Szepesvári, C.
\newblock X-armed bandits.
\newblock \emph{Journal of Machine Learning Research}, 2011.

\bibitem[Castro et~al.(2005)Castro, Willett, and Nowak]{Active_Castro2005}
Castro, R.~M., Willett, R.~M., and Nowak, R.~D.
\newblock Faster rates in regression via active learning.
\newblock In \emph{Neural Information Processing Systems}, 2005.

\bibitem[Chen \& Gupta(2012)Chen and Gupta]{ChenGuptaCP_Book}
Chen, J. and Gupta, A.~K.
\newblock \emph{{Parametric Statistical Change Point Analysis: With Applications to Genetics, Medicine, and Finance; 2nd ed.}}
\newblock Springer, 2012.

\bibitem[Chen et~al.(2014)Chen, Lin, King, Lyu, and Chen]{chen_combinatorial}
Chen, S., Lin, T., King, I., Lyu, M.~R., and Chen, W.
\newblock Combinatorial pure exploration of multi-armed bandits.
\newblock In \emph{Proceedings of the 28th International Conference on Neural Information Processing Systems - Volume 1}, 2014.

\bibitem[Chen et~al.(2019)Chen, Lee, Luo, and Wei]{non_stationary_CP_chen}
Chen, Y., Lee, C.-W., Luo, H., and Wei, C.-Y.
\newblock A new algorithm for non-stationary contextual bandits: Efficient, optimal and parameter-free.
\newblock In \emph{Proceedings of the Thirty-Second Conference on Learning Theory}, 2019.

\bibitem[Chen et~al.(2025)Chen, Karthik, Chee, and Tan]{chen2025optimalmultiobjectivebestarm}
Chen, Z., Karthik, P., Chee, Y.~M., and Tan, V.~Y.
\newblock Optimal multi-objective best arm identification with fixed confidence.
\newblock In \emph{The 28th International Conference on Artificial Intelligence and Statistics}, 2025.

\bibitem[Degenne \& Koolen(2019)Degenne and Koolen]{Degenne2019PureEMultipleCorrectAnswers}
Degenne, R. and Koolen, W.~M.
\newblock Pure exploration with multiple correct answers.
\newblock In \emph{Neural Information Processing Systems}, 2019.

\bibitem[Degenne et~al.(2019)Degenne, Koolen, and M{\'e}nard]{Degenne2019NonAsymptoticPE}
Degenne, R., Koolen, W.~M., and M{\'e}nard, P.
\newblock Non-asymptotic pure exploration by solving games.
\newblock In \emph{Neural Information Processing Systems}, 2019.

\bibitem[Fryzlewicz(2014)]{Fryzlewicz_2014_wild_binary_segmentation}
Fryzlewicz, P.
\newblock Wild binary segmentation for multiple change-point detection.
\newblock \emph{The Annals of Statistics}, 42, December 2014.

\bibitem[Garivier \& Kaufmann(2016)Garivier and Kaufmann]{kaufmann_16a}
Garivier, A. and Kaufmann, E.
\newblock Optimal best arm identification with fixed confidence.
\newblock In \emph{29th Annual Conference on Learning Theory}, 2016.

\bibitem[Garivier \& Moulines(2011)Garivier and Moulines]{Discounted_UCB_non_stationary}
Garivier, A. and Moulines, E.
\newblock On upper-confidence bound policies for switching bandit problems.
\newblock In \emph{Algorithmic Learning Theory}, 2011.

\bibitem[Garivier et~al.(2016)Garivier, Ménard, and Stoltz]{Garivier_chain_rule_trick}
Garivier, A., Ménard, P., and Stoltz, G.
\newblock Explore first, exploit next: The true shape of regret in bandit problems.
\newblock \emph{Mathematics of Operations Research}, 2016.

\bibitem[Garivier et~al.(2018)Garivier, Ménard, Rossi, and Menard]{garivier2018thresholdingbanditdoserangingimpact}
Garivier, A., Ménard, P., Rossi, L., and Menard, P.
\newblock Thresholding bandit for dose-ranging: The impact of monotonicity, 2018.
\newblock URL \url{https://arxiv.org/abs/1711.04454}.

\bibitem[Gopalan et~al.(2021)Gopalan, Lakshminarayanan, and Saligrama]{banditsQuickestCDP}
Gopalan, A., Lakshminarayanan, B., and Saligrama, V.
\newblock Bandit quickest changepoint detection.
\newblock \emph{Advances in Neural Information Processing Systems 34 (NeurIPS 2021)}, 2021.

\bibitem[Gramacy \& Lee(2008)Gramacy and Lee]{AdaptiveDesignSupercomputer}
Gramacy, R. and Lee, H.
\newblock Adaptive design and analysis of supercomputer experiments.
\newblock \emph{Technometrics}, 51, 2008.

\bibitem[Hall \& Molchanov(2003)Hall and Molchanov]{active_Hall2003}
Hall, P. and Molchanov, I.
\newblock {Sequential methods for design-adaptive estimation of discontinuities in regression curves and surfaces}.
\newblock \emph{The Annals of Statistics}, 2003.

\bibitem[Hayashi et~al.(2019)Hayashi, Kawahara, and Kashima]{ACPD}
Hayashi, S., Kawahara, Y., and Kashima, H.
\newblock Active change-point detection.
\newblock In \emph{Proceedings of The Eleventh Asian Conference on Machine Learning}, 2019.

\bibitem[Hou et~al.(2024)Hou, Tan, and Zhong]{hou2024almost}
Hou, Y., Tan, V. Y.~F., and Zhong, Z.
\newblock Almost minimax optimal best arm identification in piecewise stationary linear bandits.
\newblock In \emph{The Thirty-eighth Annual Conference on Neural Information Processing Systems}, 2024.

\bibitem[Jamieson \& Nowak(2014)Jamieson and Nowak]{fixed_confidence_survey}
Jamieson, K. and Nowak, R.
\newblock Best-arm identification algorithms for multi-armed bandits in the fixed confidence setting.
\newblock In \emph{2014 48th Annual Conference on Information Sciences and Systems (CISS)}, 2014.

\bibitem[Jamieson et~al.(2014)Jamieson, Malloy, Nowak, and Bubeck]{pmlr-v35-jamieson14_lilucb}
Jamieson, K., Malloy, M., Nowak, R., and Bubeck, S.
\newblock lil' ucb : An optimal exploration algorithm for multi-armed bandits.
\newblock In \emph{Proceedings of The 27th Conference on Learning Theory}, 2014.

\bibitem[Jourdan et~al.(2022)Jourdan, Degenne, Baudry, de~Heide, and Kaufmann]{TopTwoJourdan}
Jourdan, M., Degenne, R., Baudry, D., de~Heide, R., and Kaufmann, E.
\newblock Top two algorithms revisited.
\newblock \emph{Advances in Neural Information Processing Systems}, 2022.

\bibitem[Juneja \& Krishnasamy(2018)Juneja and Krishnasamy]{Juneja2018SampleCO_TAS_GENERALISATION}
Juneja, S. and Krishnasamy, S.
\newblock Sample complexity of partition identification using multi-armed bandits.
\newblock In \emph{Annual Conference Computational Learning Theory}, 2018.

\bibitem[Kaufmann \& Koolen(2021)Kaufmann and Koolen]{kaufmann_martingales}
Kaufmann, E. and Koolen, W.~M.
\newblock {Mixture Martingales Revisited with Applications to Sequential Tests and Confidence Intervals}.
\newblock \emph{{Journal of Machine Learning Research}}, 2021.

\bibitem[Kleinberg et~al.(2008)Kleinberg, Slivkins, and Upfal]{ZoomingAlgo}
Kleinberg, R., Slivkins, A., and Upfal, E.
\newblock Multi-armed bandits in metric spaces.
\newblock In \emph{Proceedings of the Fortieth Annual ACM Symposium on Theory of Computing}, 2008.

\bibitem[Lan et~al.(2009)Lan, Banerjee, and Michailidis]{Active_Lan_2009}
Lan, Y., Banerjee, M., and Michailidis, G.
\newblock Change-point estimation under adaptive sampling.
\newblock \emph{The Annals of Statistics}, 2009.

\bibitem[Lattimore \& Szepesvári(2020)Lattimore and Szepesvári]{BanditAlgosBook}
Lattimore, T. and Szepesvári, C.
\newblock \emph{Bandit Algorithms}.
\newblock Cambridge University Press, 2020.

\bibitem[Lazzaro \& Pike-Burke(2025)Lazzaro and Pike-Burke]{lazzaro2025fixedbudgetchangepointidentification}
Lazzaro, J. and Pike-Burke, C.
\newblock Fixed-budget change point identification in piecewise constant bandits.
\newblock In \emph{The 28th International Conference on Artificial Intelligence and Statistics}, 2025.

\bibitem[Magureanu et~al.(2014)Magureanu, Combes, and Proutiere]{pmlr-v35-magureanu14}
Magureanu, S., Combes, R., and Proutiere, A.
\newblock Lipschitz bandits: Regret lower bound and optimal algorithms.
\newblock In \emph{Proceedings of The 27th Conference on Learning Theory}, 2014.

\bibitem[Park et~al.(2021)Park, Qiu, Carpena-Núñez, Rao, Susner, and Maruyama]{park2021sequentialAdaptiveDesignForJumpRegression}
Park, C., Qiu, P., Carpena-Núñez, J., Rao, R., Susner, M., and Maruyama, B.
\newblock Sequential adaptive design for jump regression estimation.
\newblock In \emph{ArXiv Preprint ArXiv:1904.01648}, 2021.

\bibitem[Park et~al.(2023)Park, Waelder, Kang, Maruyama, Hong, and Gramacy]{ActiveLearningPiecewiseGP}
Park, C., Waelder, R., Kang, B., Maruyama, B., Hong, S., and Gramacy, R.
\newblock Active learning of piecewise gaussian process surrogates.
\newblock In \emph{ArXiv Preprint ArXiv:2301.08789}, 2023.

\bibitem[Qin et~al.(2017)Qin, Klabjan, and Russo]{Qin_TopTwoEI}
Qin, C., Klabjan, D., and Russo, D.
\newblock Improving the expected improvement algorithm.
\newblock In \emph{Proceedings of the 31st International Conference on Neural Information Processing Systems}, 2017.

\bibitem[Russo(2016)]{TopTwoRusso}
Russo, D.
\newblock Simple bayesian algorithms for best arm identification.
\newblock In \emph{29th Annual Conference on Learning Theory}, 2016.

\bibitem[Scott \& Knott(1974)Scott and Knott]{original_binary_segmentation_paper}
Scott, A.~J. and Knott, M.
\newblock A cluster analysis method for grouping means in the analysis of variance.
\newblock \emph{Biometrics}, \penalty0 (3):\penalty0 507--512, 1974.

\bibitem[Srinivas et~al.(2010)Srinivas, Krause, Kakade, and Seeger]{GPUCB_paper}
Srinivas, N., Krause, A., Kakade, S., and Seeger, M.
\newblock Gaussian process optimization in the bandit setting: No regret and experimental design.
\newblock In \emph{ICML 2010 - Proceedings, 27th International Conference on Machine Learning}, 2010.

\bibitem[Thuot et~al.(2024)Thuot, Carpentier, Giraud, and Verzelen]{thuot2024activeclusteringbanditfeedback}
Thuot, V., Carpentier, A., Giraud, C., and Verzelen, N.
\newblock Active clustering with bandit feedback, 2024.
\newblock URL \url{https://arxiv.org/abs/2406.11485}.

\bibitem[Verzelen et~al.(2020)Verzelen, Fromont, Lerasle, and Reynaud-Bouret]{Verzelen2020OptimalChangeDetection_localization}
Verzelen, N., Fromont, M., Lerasle, M., and Reynaud-Bouret, P.
\newblock Optimal change-point detection and localization.
\newblock \emph{The Annals of Statistics}, 2020.

\bibitem[Yang et~al.(2022)Yang, Zhong, and Tan]{Yang2022OptimalClustering}
Yang, J., Zhong, Z., and Tan, V. Y.~F.
\newblock Optimal clustering with bandit feedback.
\newblock \emph{J. Mach. Learn. Res.}, 2022.

\bibitem[Yavas et~al.(2025)Yavas, Huang, Tan, and Scarlett]{yavas2024generalframeworkclusteringdistribution}
Yavas, R.~C., Huang, Y., Tan, V.~Y., and Scarlett, J.
\newblock A general framework for clustering and distribution matching with bandit feedback.
\newblock \emph{IEEE Transactions on Information Theory}, 2025.

\end{thebibliography}
\bibliographystyle{icml2025}

%%%%%%%%%%%%%%%%%%%%%%%%%%%%%%%%%%%%%%%%%%%%%%%%%%%%%%%%%%%%%%%%%%%%%%%%%%%%%%%
%%%%%%%%%%%%%%%%%%%%%%%%%%%%%%%%%%%%%%%%%%%%%%%%%%%%%%%%%%%%%%%%%%%%%%%%%%%%%%%
% APPENDIX
%%%%%%%%%%%%%%%%%%%%%%%%%%%%%%%%%%%%%%%%%%%%%%%%%%%%%%%%%%%%%%%%%%%%%%%%%%%%%%%
%%%%%%%%%%%%%%%%%%%%%%%%%%%%%%%%%%%%%%%%%%%%%%%%%%%%%%%%%%%%%%%%%%%%%%%%%%%%%%%
%%%%%%%%%%%%%%%%%%%%%%%%%%%%%%%%%%%%%%%%%%%%%%%%%%%%%%%%%%%%%%%%%%%%%%%%%%%%%%%
%%%%%%%%%%%%%%%%%%%%%%%%%%%%%%%%%%%%%%%%%%%%%%%%%%%%%%%%%%%%%%%%%%%%%%%%%%%%%%%
% APPENDIX
%%%%%%%%%%%%%%%%%%%%%%%%%%%%%%%%%%%%%%%%%%%%%%%%%%%%%%%%%%%%%%%%%%%%%%%%%%%%%%%
%%%%%%%%%%%%%%%%%%%%%%%%%%%%%%%%%%%%%%%%%%%%%%%%%%%%%%%%%%%%%%%%%%%%%%%%%%%%%%%
%%%%%%%%%%%%%%%%%%%%%%%%%%%%%%%%%%%%%%%%%%%%%%%%%%%%%%%%%%%%%%%%%%%%%%%%%%%%%%%
%%%%%%%%%%%%%%%%%%%%%%%%%%%%%%%%%%%%%%%%%%%%%%%%%%%%%%%%%%%%%%%%%%%%%%%%%%%%%%%
% APPENDIX
%%%%%%%%%%%%%%%%%%%%%%%%%%%%%%%%%%%%%%%%%%%%%%%%%%%%%%%%%%%%%%%%%%%%%%%%%%%%%%%
%%%%%%%%%%%%%%%%%%%%%%%%%%%%%%%%%%%%%%%%%%%%%%%%%%%%%%%%%%%%%%%%%%%%%%%%%%%%%%%

\newpage
\appendix
\onecolumn
\section*{Appendix}
% \tableofcontents
%%%%%%%%%%%%%%%%%%%%%%%%%%%%%%%%%%%%%%%%%%%%%%%%%%%%%%%%%%%%%%%%%%%%%%%%%%%%%%%
%%%%%%%%%%%%%%%%%%%%%%%%%%%%%%%%%%%%%%%%%%%%%%%%%%%%%%%%%%%%%%%%%%%%%%%%%%%%%%%

% ADDITIONAL SIMS
%%%%%%%%%%%%%%%%%%%%%%%%%%%%%%%%%%%%%%%%%%%%%%%%%%%%%%%%%%%%%%%%%%%%%%%%%%%%%%%
%%%%%%%%%%%%%%%%%%%%%%%%%%%%%%%%%%%%%%%%%%%%%%%%%%%%%%%%%%%%%%%%%%%%%%%%%%%%%%%
\section{Additional Experiments}\label{app_sims}

\begin{figure}[ht]
\vskip 0.2in
\begin{center}
\centerline{\includegraphics[width=0.5\columnwidth]{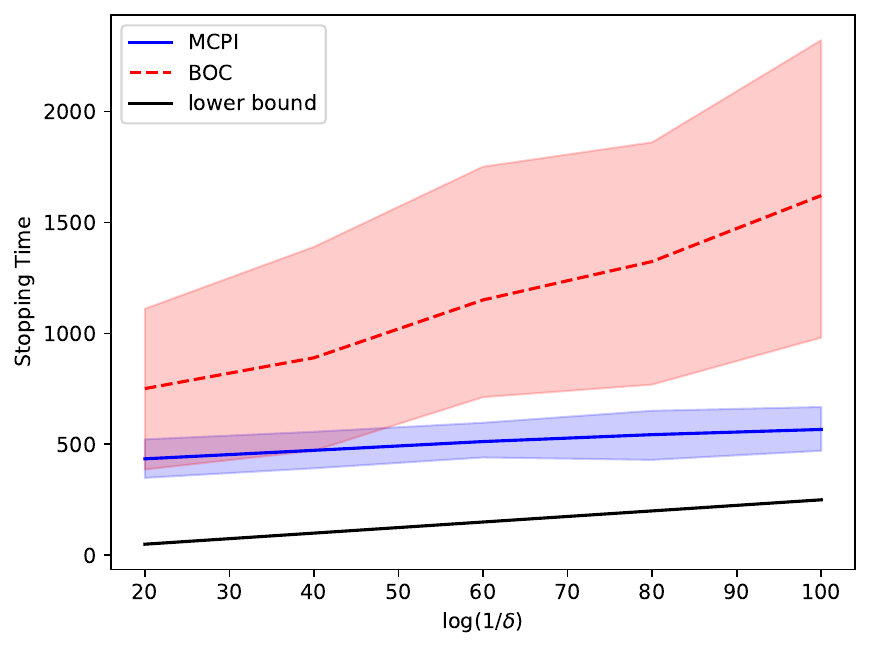}}
\caption{We run the MCPI and BOC algorithm at a range of values for $\delta$ in environment $v_2$. At each confidence level we repeat 100 runs and plot the average stopping time with 90 percent confidence intervals. We also plot the lower bound in Theorem \ref{theorem_any_m_lb_general_case}.}
\label{fig2}
\end{center}
\vskip -0.4in
\end{figure}

\begin{figure}[ht]
\vskip 0.2in
\begin{center}
\centerline{\includegraphics[width=0.5\columnwidth]{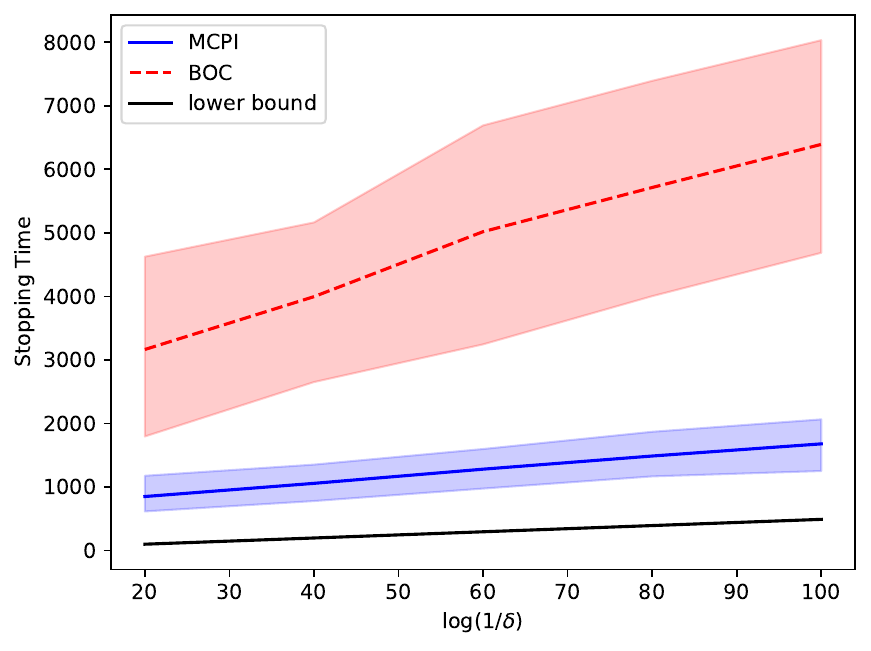}}
\caption{We run the MCPI and BOC algorithm at a range of values for $\delta$ in environment $v_3$. At each confidence level we repeat 100 runs and plot the average stopping time with 90 percent confidence intervals. We also plot the lower bound in Theorem \ref{theorem_any_m_lb_general_case}.}
\label{fig3}
\end{center}
\vskip -0.1in
\end{figure}

In addition to the experiments shown in Section \ref{section:experiments}, we simulate our methods in environments with more than one change point. For example, in Figure \ref{fig2} we consider and environment $v_2$ with $K=19$ actions and $N=2$ change points with mean rewards
$\mu_{v_2} = (2,2,2,2,2,2,4,4,4,4,4,4,4,0,0,0,0,0,0)$. In Figure \ref{fig2} we again plot the average stopping time when running the MPCI algorithm inputting $N=2$ 100 times at different values for $\log(1/\delta)$. We also run BOC with oracle weights. We again see that the stopping time for BOC increases at a much faster rate than MCPI (which stays approximately parallel to the lower bound from Theorem \ref{theorem_any_m_lb_general_case}). We see similar results when running the same experiment but using environment $v_3$ $K=9$ actions and with $N=3$ change points. Here the mean rewards are $\mu_{v_3} = (2,2,3,3,3,3,1,1,4)$. The results from simulations on this environment are in Figure \ref{fig3}.

\newpage
Finally, we also run MCPI in an environment $v_4$ with $m=5$ change points $\mu_{v_4} = (2,2,2.5,2.5,3,3, 2, 2, 1.5, 1.5,1.5,1.5,1.25,1.25)$. In this case, however we only input $N=1$ into MCPI such that MCPI will search for only one change point. In this case, where there are an additional 4 change points present, we simulate the performance of MCPI 1000 ties at each of a range of $\delta$ values. We then plot our average observed stopping time and compare this with the lower bound in Theorem \ref{theorem_any_m_lb_general_case} with $N=1$. We again see that the average stopping time of MCPI is approximately parallel to the lower bound, supporting our theoretical results that MCPI is an asymptotically optimal Any-$(N,\delta)$ policy which is robust to an unknown number of additional change points present in the environment.

\begin{figure}[H]
\vskip 0.2in
\begin{center}
\centerline{\includegraphics[width=0.5\columnwidth]{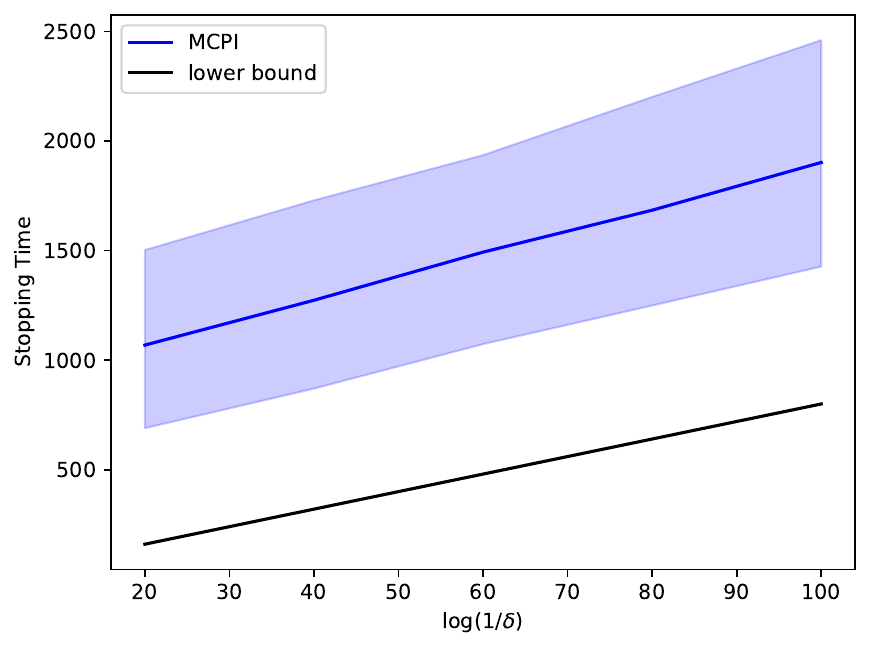}}
\caption{We run the MCPI algorithm with $N=1$ at a range of values for $\delta$ in environment $v_4$. At each confidence level we repeat 100 runs and plot the average stopping time with 90 percent confidence intervals. We also plot the lower bound in Theorem \ref{theorem_any_m_lb_general_case}.}
\label{fig4}
\end{center}
\vskip -0.1in
\end{figure}

%%%%%%%%%%%%%%%%%%%%%%%%%%%%%%%%%%%%%%%%%%%%%%%%%%%%%%%%%%%%%%%%%%%%%%%%%%%%%%%
%%%%%%%%%%%%%%%%%%%%%%%%%%%%%%%%%%%%%%%%%%%%%%%%%%%%%%%%%%%%%%%%%%%%%%%%%%%%%%%
% PROOFS
%%%%%%%%%%%%%%%%%%%%%%%%%%%%%%%%%%%%%%%%%%%%%%%%%%%%%%%%%%%%%%%%%%%%%%%%%%%%%%%
%%%%%%%%%%%%%%%%%%%%%%%%%%%%%%%%%%%%%%%%%%%%%%%%%%%%%%%%%%%%%%%%%%%%%%%%%%%%%%%
\newpage
\section*{Proofs for Section \ref{section:one_cp}}
\section{Proof Sketch of Theorem \ref{theorem_1cp_general}} \label{app_proof_theorem_1cp_general}

The proof of this Theorem can be seen in both Theorem 1 of \citep{kaufmann_16a} and Theorem 33.5 of \citep{BanditAlgosBook}, where the difference is that we include additional structural assumptions on the mean rewards in our environments and we have a different objective from best arm identification. In particular, we reiterate the infimum in Theorem \ref{theorem_1cp_general} is over the `alternative' set of piecewise constant environments $v'$ which have exactly one change point which is not equal to $x^*_{v}$.

We repeat the two important steps here for clarity and motivation. Firstly, using a change of measure argument and considering any $v' \in V^{alt}_{K,1}(x^*_v)$, they show that 
    \begin{equation} \label{existing_lower_bound_proof_eqn1}
        \sum_{i=1}^K \mathbb{E}_{\pi,v}[T_i(t)] D(v_i,v_i') \geq \log(1/4\delta)
    \end{equation}
    Secondly, they use \eqref{existing_lower_bound_proof_eqn1} and definition of $c^*(v)$ to show the following steps
    \begin{align}
        \frac{\mathbb{E}_{\pi,v}[\tau_\delta]}{c^*(v)} &= \mathbb{E}_{\pi,v}[\tau_\delta] \sup_{\alpha \in \mathcal{P}_{K}}\inf_{v' \in V^{alt}(x^*_v)} \sum_{i=1}^K \alpha_i D(v_i,v_i') \nonumber\\
        &\geq \mathbb{E}_{\pi,v}[\tau_\delta] \inf_{v' \in V^{alt}(x^*_v)} \sum_{i=1}^K \frac{\mathbb{E}_{\pi,v}[T_i(\tau_\delta)]}{\mathbb{E}_{\pi,v}[\tau_\delta]} D(v_i,v_i') \label{optimal_proportions_intuition} \\
        &= \inf_{v' \in V^{alt}(x^*_v)} \sum_{i=1}^K \mathbb{E}_{\pi,v}[T_i(\tau_\delta)]D(v_i,v_i') \nonumber\\
        &\geq \log(1/4\delta)\nonumber
    \end{align}

    The proof is then complete. \qed
    
Recall the definition of $\alpha^*$ as the solution of the optimisation problem in equation \eqref{alpha_star_opt_problem_1cp}. Then we observe that the only way to achieve equality in equation \eqref{optimal_proportions_intuition} is for  $\alpha^*_i = \frac{\mathbb{E}_{\pi,v}[T_i(\tau_\delta)]}{\mathbb{E}_{\pi,v}[\tau_\delta]}$ to hold. Hence, to asymptotically achieve the lower bound presented in Theorem \ref{theorem_1cp_general}, we want to have the proportion of times we play each action equal to $\alpha^*$. Hence, we often refer to $\alpha^*$ as the vector of optimal proportions.

%%%%%%%%%%%%%%%%%%%%%%%%%%%%%%%%%%%%%%%%%%%%%%%%%%%%%%%%%%%%%%%%%%%%%%%%%%%%%%%
%%%%%%%%%%%%%%%%%%%%%%%%%%%%%%%%%%%%%%%%%%%%%%%%%%%%%%%%%%%%%%%%%%%%%%%%%%%%%%%
%%%%%%%%%%%%%%%%%%%%%%%%%%%%%%%%%%%%%%%%%%%%%%%%%%%%%%%%%%%%%%%%%%%%%%%%%%%%%%%
%%%%%%%%%%%%%%%%%%%%%%%%%%%%%%%%%%%%%%%%%%%%%%%%%%%%%%%%%%%%%%%%%%%%%%%%%%%%%%%
\newpage
\section{Proof of Theorem \ref{theorem_exact_1cp_lb_explicit}}\label{app_proof_theorem_exact_1cp_lb_explicit}
We show that we can explicitly solve the following optimization problem for $c^*(v)^{-1}$.

    \begin{equation*}
        c^*(v)^{-1} =  \sup_{\alpha \in \mathcal{P}_{K}}\inf_{v' \in V^{alt}_{K,1}(x^*_v)} \sum_{i=1}^K \alpha_i D\left(v(i),v'(i)\right).
    \end{equation*}

    First, suppose we fix some $\alpha \in \mathcal{P}_{K}$. Then we try to to choose a $v' \in V^{alt}_{K,1}(x^*_v)$ to minimize
     \begin{equation}\label{lower_bound_alpha_calculation_sum}
        \sum_{i=1}^K \alpha_i D(v_i,v_i').
    \end{equation}

  Denote the mean rewards in environment $v$ as $(\mu_i)_{i=1}^K$ and the mean rewards in environment $v$ as $(\mu_i')_{i=1}^K$.

    \textbf{Case 1:} Consider some $v'$ such that $x^*_{v'}> x^*_v$. Then, we can rewrite the sum in \eqref{lower_bound_alpha_calculation_sum} as

    \begin{align}
        \sum_{i=1}^K \alpha_i D(v_i,v_i') &= \sum_{i=1}^{x^*_{v}} \alpha_i D(v_i,v_i') + \sum_{i=x^*_v+1}^{x^*_{v'}} \alpha_i D(v_i,v_i') + \sum_{i=x^*_{v'}+1}^{K} \alpha_i D(v_i,v_i')\nonumber\\ 
        &=\frac{1}{2\sigma^2}\left(\sum_{i=1}^{x^*_{v}} \alpha_i(\mu_1-\mu_1')^2+ \sum_{i=x^*_v+1}^{x^*_{v'}} \alpha_i(\mu_K-\mu_1')^2 + \sum_{i=x^*_{v'}+1}^{K} \alpha_i(\mu_K-\mu_K')^2\right) \label{lower_bound_minimisation_decomposition}
    \end{align}
    
    Where \eqref{lower_bound_minimisation_decomposition} comes from the definition of the KL divergence between two Gaussian distributions. In order to minimize the third term in \eqref{lower_bound_minimisation_decomposition}, we  choose $\mu_K'=\mu_K$. In order to minimize the first two terms, we choose 

    \begin{equation*}
        \mu_1' = \mu_1 + (\mu_K-\mu_1)\frac{\sum_{i=x^*_v+1}^{x^*_{v'}} \alpha_i}{\sum_{i=1}^{x^*_{v}} \alpha_i+\sum_{i=x^*_v+1}^{x^*_{v'}} \alpha_i}
    \end{equation*}

    We plug these choices for $\mu_1',\mu_K'$ into \eqref{lower_bound_minimisation_decomposition} to get

    \begin{equation}\label{eqn_app_case1_minsum}
         \sum_{i=1}^K \alpha_i D(v_i,v_i') =  \frac{\Delta^2\left(\sum_{i=1}^{x^*_{v}} \alpha_i \right) \left(\sum_{i=x^*_v+1}^{x^*_{v'}} \alpha_i\right)}{2\sigma^2\left(\sum_{i=1}^{x^*_{v}} \alpha_i+\sum_{i=x^*_v+1}^{x^*_{v'}} \alpha_i\right)}
    \end{equation}

    which is minimised when $x^*_{v'}= x^*_v+1$.\textcolor{red}\\

    \textbf{Case 2:} Now, consider some $v'$ such that $x^*_{v'} < x^*_v$. Similar to case 1, we can minimise the sum in \eqref{lower_bound_alpha_calculation_sum} by setting

    \begin{align*}
        \mu_1'&=\mu_1\\
        \mu_K' &= \mu_1 +(\mu_K-\mu_1)\frac{\sum_{i=x^*_{v'}+1}^{x^*_{v}} \alpha_i}{\sum_{i=x^*_{v'}+1}^{x^*_{v}} \alpha_i+\sum_{i=x^*_v+1}^{K} \alpha_i}\\
        x^*_{v'} &= x^*_{v} - 1
    \end{align*}

    to attain 

    \begin{equation}\label{eqn_app_case2_minsum}
         \sum_{i=1}^K \alpha_i D(v_i,v_i') =  \frac{\Delta^2\left(\sum_{i=x^*_{v'}+1}^{x^*_{v}} \alpha_i\right) \left(\sum_{i=x^*_v+1}^{K} \alpha_i\right)}{2\sigma^2\left(\sum_{i=x^*_{v'}+1}^{x^*_{v}} \alpha_i+\sum_{i=x^*_v+1}^{K} \alpha_i\right)}
    \end{equation}

    \textbf{Combining Cases To Find $\alpha^*$:} The above arguments show that the `closest' environments to $v$ are when we shift the change point by 1 to either the left or the right. We therefore have that the unique maximizing choice for $\alpha^*$ will minimize the minimum of \eqref{eqn_app_case1_minsum} and \eqref{eqn_app_case2_minsum}.

    \begin{align*}
        \argmax_{\alpha \in \mathcal{P}_{K}} \inf_{v' \in V^{alt}(x^*_v)} \sum_{i=1}^k \alpha_i D(v_i,v_i') 
        &= \argmax_{\alpha \in \mathcal{P}_{K-1}}\,\min\left\{\frac{\Delta^2\left(\sum_{i=1}^{x^*_{v}} \alpha_i \right) \left( \alpha_{x^*_v+1}\right)}{2\sigma^2\left(\sum_{i=1}^{x^*_{v}} \alpha_i+\alpha_{x^*_v+1}\right)}, \frac{\Delta^2\left( \alpha_{x^*_v}\right) \left(\sum_{i=x^*_v+1}^{K} \alpha_i\right)}{2\sigma^2\left(\alpha_{x^*_v}+\sum_{i=x^*_v+1}^{K} \alpha_i\right)}\right\}\\
        &= \begin{cases}
   1/2       & \text{if } i \in \{x^*_v, x^*_v+1\} \\
   0      & \text{otherwise}
  \end{cases}
    \end{align*}
    Hence, we have attained \eqref{eqn_1cp_alpha_star}. This also proves the Theorem by plugging this back into \eqref{lower_bound_alpha_calculation_sum} and subsequently  our calculation for $c^*(v)$ into Theorem \ref{theorem_1cp_general}. \qed

%%%%%%%%%%%%%%%%%%%%%%%%%%%%%%%%%%%%%%%%%%%%%%%%%%%%%%%%%%%%%%%%%%%%%%%%%%%%%%%
%%%%%%%%%%%%%%%%%%%%%%%%%%%%%%%%%%%%%%%%%%%%%%%%%%%%%%%%%%%%%%%%%%%%%%%%%%%%%%%
%%%%%%%%%%%%%%%%%%%%%%%%%%%%%%%%%%%%%%%%%%%%%%%%%%%%%%%%%%%%%%%%%%%%%%%%%%%%%%%
%%%%%%%%%%%%%%%%%%%%%%%%%%%%%%%%%%%%%%%%%%%%%%%%%%%%%%%%%%%%%%%%%%%%%%%%%%%%%%%
\newpage
\section{Proof of Proposition \ref{prop_stopping_time}}\label{app_proof_prop_stopping_time}

Before starting the proof, we note that in previous works using Track-and-Stop methods,
%\citep[such as][]{Degenne2019NonAsymptoticPE}
a Chernoff stopping time was used \citep{kaufmann_16a}. This stopping time stops the algorithm when an infimum of a self-normalized sum \citep{kaufmann_martingales,BanditAlgosBook} crosses some threshold $\beta$. The general form of these stopping times are
\begin{equation}\label{eqn_stopping_time_general_form}
    \min\left\{t:\inf_{v' \in V^{alt}_{K,1}(\hat{x})} \sum_{i=1}^K T_i(t) D\left(v'(i),\hat{v}(i)\right) \geq \beta(t,\delta)\right\}.
\end{equation}
where $\hat{v}$ is an estimate for the environment.
However, in our setting since it is optimal to only play either side of the change point, we demonstrate that it is sufficient to only consider the self-normalized sum over actions $ i \in \{\hat{x}_t,\hat{x}_t+1\}$ since we expect the contribution from these terms to dominate. Namely, we will define our stopping time to be 

\begin{equation}\label{eqn_stopping_time_general_form_only2sum}
    \tau_\delta = \min\left\{t:\inf_{v' \in V^{alt}_{K,1}(\hat{x})} \sum_{i=\hat{x}_t}^{\hat{x}_t+1} T_i(t) D\left(v'(i),\hat{v}(i)\right) \geq \beta(t,\delta)\right\}.
\end{equation}

We now prove that \eqref{eqn_stopping_time_general_form_only2sum} is an appropriate stopping time. Fix an environment $v\in V_{K,1}$ and a policy $\pi$ with the stated stopping time \eqref{eqn_stopping_time_general_form_only2sum}. Denote $\mu \in \mathbb{R}^K$ as the mean reward vector in environment $v$. Use any estimate for the change point. We then have the following sequence of inequalities, where the first steps are motivated by the Track-and-Stop analysis for best arm identification in \citet{BanditAlgosBook}.
    \begin{align}
        \mathbb{P}_{\pi,v}(\hat{x}_{\tau} \neq x^*_v)  &= \mathbb{P}_{\pi,v}(v \in V^{alt}(\hat{x}_{\tau}))\nonumber\\
        &\leq \mathbb{P}_{\pi,v}\left(\frac{1}{2}\left[T_{\hat{x}_\tau}(\tau)(\hat{\mu}_{\hat{x}_\tau}(\tau) - \mu_{\hat{x}_\tau})^2 + T_{\hat{x}_\tau+1}(\tau)(\hat{\mu}_{\hat{x}_\tau+1}(\tau) - \mu_{\hat{x}_\tau+1})^2\right] \geq \beta(\tau,\delta)\right) \label{stopping_by_def}\\
        &\leq \mathbb{P}_{\pi,v}\left(\exists s\in\mathbb{N}^+, \exists k \in [K-1] :  \frac{1}{2}\left[T_{k}(s)(\hat{\mu}_{k}(s) - \mu_{k})^2 + T_{k+1}(s)(\hat{\mu}_{k+1}(s) - \mu_{k+1})^2\right] \geq \beta(s,\delta)\right)\nonumber\\
        &\leq \sum_{k=1}^{K-1}\sum_{s=1}^\infty\mathbb{P}_{\pi,v}\left(\frac{1}{2}\left[T_{k}(s)(\hat{\mu}_{k}(s) - \mu_{k})^2 + T_{k+1}(s)(\hat{\mu}_{k+1}(s) - \mu_{k+1})^2\right] \geq \beta(s,\delta)\right)\nonumber\\
        &\leq \sum_{k=1}^{K-1}\sum_{s=1}^\infty\exp(3) \left(\frac{\beta(s,\delta)\left(\beta(s,\delta)\log(s)+1\right)}{2}\right)^2 \exp(-\beta(s,\delta))\label{stopping_existing_trick}
    \end{align}

Where \eqref{stopping_by_def} comes form the definition of our stopping time \eqref{eqn_stopping_time_general_form_only2sum} and the final inequality \eqref{stopping_existing_trick} comes from Theorem 2 of \cite{pmlr-v35-magureanu14}. Now, we will make similar arguments to \citet{garivier2018thresholdingbanditdoserangingimpact} to complete the proof. We first point out some helpful observations.

\begin{lemma} \label{stopping_useful_facts}The following inequalities hold for all $t \geq 1$ and $\delta \in (0,1)$
    \begin{enumerate}[(i)]
    \item $\beta(t,\delta) \leq 9 \log\left(\frac{t\gamma(K-1)}{\delta}\right)$
    \item $\log(t) \leq \beta(t,\delta)$
    \item $\beta(t,\delta) \geq 1$
\end{enumerate}
\end{lemma}
\begin{proof}
     These results follow from the definition of our threshold $\beta$, defined in equation \eqref{eqn_beta_definition} and the fact that $\gamma(K-1)/\delta \geq 3$.
\end{proof}
Now, by plugging in our definition for $\beta(t,\delta)$ into equation
\eqref{stopping_existing_trick}, we get the following sequence of inequalities.
    \begin{align}
        \mathbb{P}_{\pi,v}(\hat{x}_{\tau_\delta} \neq x^*_v)  &\leq  \sum_{k=1}^{K-1}\sum_{s=1}^\infty \frac{\exp(3)\delta}{4s\gamma(K-1)} \frac{\left(\beta(s,\delta)\left(\beta(s,\delta)\log(s)+1\right)\right)^2}{\log^8(s\gamma(K-1)/\delta)}\nonumber\\
        &\leq  \sum_{k=1}^{K-1}\sum_{s=1}^\infty \frac{\exp(3)\delta}{4s\gamma(K-1)} \frac{4\left(\beta(s,\delta)\right)^6}{\log^8(s\gamma(K-1)/\delta)} \label{stopping_using_facts1}\\
        &\leq  \sum_{k=1}^{K-1}\sum_{s=1}^\infty \frac{\exp(3)\delta}{4s\gamma(K-1)} \frac{4\left(9 \log\left(\frac{t\gamma(K-1)}{\delta}\right)\right)^6}{\log^8(s\gamma(K-1)/\delta)} \label{stopping_using_facts2}\\
        &= \sum_{k=1}^{K-1}\sum_{s=1}^\infty \frac{\exp(3)\delta}{s\gamma(K-1)} \frac{9^6}{\log^2(s\gamma(K-1)/\delta)}\nonumber\\
        &= \sum_{k=1}^{K-1} \frac{9^6\exp(3)\delta}{\gamma(K-1)} \sum_{s=1}^\infty\frac{1}{s\log^2(s\gamma(K-1)/\delta)}\nonumber\\
        &\leq  \sum_{k=1}^{K-1} \frac{9^6\exp(3)\delta}{\gamma(K-1)} \sum_{s=1}^\infty\frac{1}{s\log^2(3s)} \label{stopping_using_facts3}\\
        &\leq  \sum_{k=1}^{K-1} \frac{9^6\exp(3)\delta}{\gamma(K-1)} \frac{2}{\log(3)} \label{stopping_using_facts4}\\
        &= \delta \nonumber
    \end{align}
Where equations \eqref{stopping_using_facts1} and \eqref{stopping_using_facts2} come from using the facts from Lemma \ref{stopping_useful_facts}. Furthermore, the inequality in \eqref{stopping_using_facts3} is true since we have  $\gamma(K-1)/\delta \geq 3$ from our definition of $\gamma$. Inequality \eqref{stopping_using_facts4} is true by using an upper bounding integral.
\begin{equation*}
    \sum_{s=1}^\infty\frac{1}{s\log^2(3s)} \leq \frac{1}{log^2(3)} + \int_{s=1}^\infty \frac{1}{s\log^2(3s)} = \frac{1}{log^2(3)} + \frac{1}{3log(3)} \leq \frac{2}{\log(3)}
\end{equation*}
The final equation comes from our definition of $\gamma$. We have therefore proven that the stopping time \eqref{eqn_stopping_time_general_form_only2sum} is an appropriate stopping time. 

Finally, by using similar steps as in the proof of Theorem \ref{theorem_exact_1cp_lb_explicit} in Appendix \ref{app_proof_theorem_exact_1cp_lb_explicit} we can prove the following lemma.

\begin{lemma} The solution to the optimization problem within stopping time is \eqref{eqn_stopping_time_general_form_only2sum} is
    $$\inf_{v' \in V^{alt}_{K,1}(\hat{x})} \sum_{i=\hat{x}_t}^{\hat{x}_t+1} T_i(t) D\left(v'(i),\hat{v}(i)\right) = \frac{T_{\hat{x}_t}(t)T_{\hat{x}_t+1}(t)}{2(T_{\hat{x}_t}(t)+T_{\hat{x}_t+1}(t))}      \hat{\Delta}^2_{\hat{x}_t}$$
\end{lemma}
By plugging this into equation \eqref{eqn_stopping_time_general_form_only2sum}, recover the form of the stopping time in Proposition \ref{prop_stopping_time} and the proof is complete.
\qed

%%%%%%%%%%%%%%%%%%%%%%%%%%%%%%%%%%%%%%%%%%%%%%%%%%%%%%%%%%%%%%%%%%%%%%%%%%%%%%%
%%%%%%%%%%%%%%%%%%%%%%%%%%%%%%%%%%%%%%%%%%%%%%%%%%%%%%%%%%%%%%%%%%%%%%%%%%%%%%%
%%%%%%%%%%%%%%%%%%%%%%%%%%%%%%%%%%%%%%%%%%%%%%%%%%%%%%%%%%%%%%%%%%%%%%%%%%%%%%%
%%%%%%%%%%%%%%%%%%%%%%%%%%%%%%%%%%%%%%%%%%%%%%%%%%%%%%%%%%%%%%%%%%%%%%%%%%%%%%%
% \section{Proof of Proposition \ref{prop_complexity_UB_HP}}\label{app_proof_prop_complexity_UB_HP}

%%%%%%%%%%%%%%%%%%%%%%%%%%%%%%%%%%%%%%%%%%%%%%%%%%%%%%%%%%%%%%%%%%%%%%%%%%%%%%%
%%%%%%%%%%%%%%%%%%%%%%%%%%%%%%%%%%%%%%%%%%%%%%%%%%%%%%%%%%%%%%%%%%%%%%%%%%%%%%%
%%%%%%%%%%%%%%%%%%%%%%%%%%%%%%%%%%%%%%%%%%%%%%%%%%%%%%%%%%%%%%%%%%%%%%%%%%%%%%%
%%%%%%%%%%%%%%%%%%%%%%%%%%%%%%%%%%%%%%%%%%%%%%%%%%%%%%%%%%%%%%%%%%%%%%%%%%%%%%%
% \section{Proof of Proposition \ref{prop_complexity_UB}}\label{app_proof_prop_complexity_UB}

%%%%%%%%%%%%%%%%%%%%%%%%%%%%%%%%%%%%%%%%%%%%%%%%%%%%%%%%%%%%%%%%%%%%%%%%%%%%%%%
%%%%%%%%%%%%%%%%%%%%%%%%%%%%%%%%%%%%%%%%%%%%%%%%%%%%%%%%%%%%%%%%%%%%%%%%%%%%%%%
%%%%%%%%%%%%%%%%%%%%%%%%%%%%%%%%%%%%%%%%%%%%%%%%%%%%%%%%%%%%%%%%%%%%%%%%%%%%%%%
%%%%%%%%%%%%%%%%%%%%%%%%%%%%%%%%%%%%%%%%%%%%%%%%%%%%%%%%%%%%%%%%%%%%%%%%%%%%%%%
\newpage
\section{Proof of Proposition \ref{TAS_asymp_UB}} \label{Appendix_proof_TAS_asymp_UB}

For this asymptotic upper bound our will stay somewhat close the structure of the original track and stop analysis from \cite{kaufmann_16a}. The following Lemma is  a consequence from Lemma 6 in \cite{Degenne2019NonAsymptoticPE}, the idea being that we define the good event $E_T$ to be when all of the empirical means for all the arms are well behaved at all times up to time $T$. 

\begin{lemma}
    Define the event
    \begin{equation}\label{good_event_final_upper_bound_asymp_proof}
        E_t = \left\{ \forall s \leq t, \forall a \in [K] \quad |\hat{\mu}_{a}(s) - \mu_{a}|^2 \leq \frac{4\log(t) + 2\log(2\log(t)) +1}{T_a(s )} \right\}
    \end{equation}
    Then 
\begin{equation*}
    \forall t \geq 3, \quad \mathbb{P}_\mu (\mathcal{E}_t^c) \leq 2eK \frac{\log t}{t^{2}}, \quad \sum_{t=3}^{+\infty} \mathbb{P}_\mu (\mathcal{E}_t^c) \leq 2eK.
\end{equation*}
and 
\[
\sum_{t=3}^{+\infty} \mathbb{P}_\mu (\mathcal{E}_t^c) \leq 2eK \sum_{t=3}^{+\infty} \frac{\log t}{t^{2}} \leq 2eK \int_{x=1}^{+\infty} \frac{\log x}{x^{2}} dx = 2eK.
\]
\end{lemma}

Now, if the good event $E_t$ holds and our empirical mean rewards for all the arms are well behaved, then we know that even our simple estimator $\hat{x}_t$ for the change point will eventually always be correct after some time $T_1$. We write this formally in the following proposition.

Now, by defining
$$T_1(v) = \min \left\{T:  \frac{4\log(T) + 2\log(2\log(T)) +1}{(\sqrt{T}-K)_+} < \Delta^2/4\right\},$$
we get the following useful proposition.

\begin{proposition}\label{prop_tracking}
    % There exists a constant $T_1 \in \mathbb{N}_+$ (which depends only on the environment) such that f
    For any $T \geq T_1$, on the event $E_T$, we have that our estimate is correct.
    $\hat{x}_T=x^*$
\end{proposition}
\begin{proof}
    Under event $E_T$ and for $T>T_1(v)$ we have that the empirical means at time $T$ are well concentrated $\forall a \in [K] \quad |\hat{\mu}_{a}(T) - \mu_{a}|^2 \leq \Delta^2/4$. This is true since our forced exploration will enforce that $\forall s \in \mathbb{N}\quad T_a(s) \geq (\sqrt{s}-K)_+$. Hence, by our definition for our simple estimator \eqref{eqn_estimator_simple_defn}, we know that the estimate for the change point at time $T$ will be correct.
\end{proof}

Hence, if we are always correct in our estimate for the change point after time $T_1(v)$, then our estimate for the optimal proportions $\hat{\alpha}$ will be correct after time $T_1(v)$. Therefore, due to our tracking procedure, the proportions of samples we make for each action will get closer to the optimal proportions at the following rate.

\begin{proposition}\label{prop_tracking2}
    There exists a constant $T_1(v) \in \mathbb{N}_+$ (which depends only on the environment) such that for any $T \geq T_1$, on the event $E_T$, we have
    $$\forall t > \sqrt{T},  \max_i |T_i(t)/t - \alpha^*_i| \leq \frac{1+(K-2)T_1(v)}{2t}$$
\end{proposition}

Now, if the proportion of plays for each action is getting closer to the optimal proportions as described in the above lemma, then we know that the number of times we play actions $x^*$ and $x^*+1$ will get closer to $t/2$ and therefore the value of $Z_t$ (defined in \eqref{eqn_Zt_definition} for our stopping time in equation \eqref{eqn_defn_stopping_time}) will get larger. Hence, we are able to provide a lower bound for $Z_t$. In order to do so, we first make the following definition.

\begin{definition}
    Let 
    \begin{align*}
        l(t) &= \frac{1+(K-2)T_1(v)}{2t}\\
        r'(t) &= \frac{4\log(t) + 2\log(2\log(t)) +1}{(\sqrt{t}-K)_+}
    \end{align*}
    Then we define $\Tilde{c}(t)$ as the following quantity.
    $$\Tilde{c}(t) := 0.5 \frac{(1/2 - l(t))^2}{1+2l(t)}(\Delta-r'(t))^2$$
    Noting that as $t \rightarrow \infty$, we have $\Tilde{c}(t) \rightarrow \frac{\Delta^2}{8} = c^*(v)^{-1}$ which was the optimal coefficient defined for the lower bound Theorem \ref{theorem_exact_1cp_lb_explicit} defined in \eqref{eqn_1cp_opt_problem}.
\end{definition}
Then, as a concequence of our true proportion of plays approaching the optimal ones, we can write the following lower bound for $Z_t$.
\begin{proposition}
    Under event $E_T(\epsilon)$ we have for all $t > \sqrt{T}$ when $T \geq 2T_1(v)$ we have
    $$Z_t \geq  t \Tilde{c}(t).$$
\end{proposition}
\begin{proof}
    Recall the definition of $Z_t$, namely
    \begin{equation*}
        Z_t = \frac{T_{\hat{x}_t}(t)T_{\hat{x}_t+1}(t)}{T_{\hat{x}_t}(t)+T_{\hat{x}_t+1}(t)}\hat{\Delta}^2.
    \end{equation*}
    where we define $\hat{\Delta}^2 := |\hat{\mu}_{\hat{x}_t}(t) - \hat{\mu}_{\hat{x}_t+1}(t)|^2$.
    The result then comes from our well behaved proportions under our well behaved event $E_T(\epsilon)$ from Proposition \ref{prop_tracking2} as well as the concentration assumptions from event $E_T(\epsilon)$ directly to lower bound $\hat{\Delta}^2$.
\end{proof}

Now, using similar steps to the original track and stop analysis of \citep{kaufmann_16a}, we have:\\

For $T \geq 2T_1(v)$ and assuming $E_T$ holds:
\begin{align*}
     \min \{\tau_\delta, T\} & \leq \sqrt{T} + \sum_{t=\sqrt{T}}^T \mathbbm{1}\{\tau_\delta>t\}\\
     & \leq \sqrt{T} + \sum_{t=\sqrt{T}}^T \mathbbm{1}\{Z_t \leq \beta(t,\delta)\}\\
     & \leq \sqrt{T} + \sum_{t=\sqrt{T}}^T \mathbbm{1}\{t\Tilde{c}(t) \leq \beta(t,\delta)\} \\
     & \leq \sqrt{T} + \sum_{t=\sqrt{T}}^T \mathbbm{1}\{t\Tilde{c}(\sqrt{T}) \leq \beta(T,\delta)\} \\
     & \leq \sqrt{T} + \frac{\beta(T,\delta)}{\Tilde{c}(\sqrt{T})}
\end{align*}

Hence, by defining the quantity
$$T_0(\delta) = \min\left\{T \in \mathbb{N}^+: \sqrt{T} + \frac{\beta(T,\delta)}{\Tilde{c}(\sqrt{T})} \leq T \right\}$$
we have that for any $T > \max\{T_0(\delta), 2T_1(v)\}$, then we have $E_T \Rightarrow \{\tau_\delta \leq T\}$, and therefore
$$\mathbb{P}(\tau_\delta >T) \leq 2eK\frac{\log(T)}{T^{2}}.$$

Subsequently, we have
\begin{align}
        \mathbb{E}[\tau] &= \sum_{t = 0}^\infty \mathbb{P}(\tau_\delta >t)\nonumber\\
        & \leq T_0(\delta) + 2T_1(v) + \sum_{t = T_0(\delta) + T_1}^\infty \mathbb{P}(\tau_\delta >t)\nonumber\\
        & \leq T_0(\delta) + 2T_1(v) + \sum_{t = T_0(\delta) + 2T_1(v)}^\infty 2eK\frac{\log(t)}{t^{2}}\nonumber\\
        & \leq T_0(\delta)+2T_1(v) + 2eK\label{eqn_1cp_asymp_proof_before_division}
    \end{align}

From this upper bound on the expected stopping time, we not that the only dependence on $\delta$ is in $T_0(\delta)$ hence we focus on studying this since the other two terms will vanish when we divide through by $\log(1/\delta)$ (asymptotically). We can now use some arguments from  \cite{garivier2018thresholdingbanditdoserangingimpact}, where we restate their useful Lemma \ref{lemma_helpful_garivier} below, and by using our own definition for the threshold $\beta$ we can carry out the following steps to upper bound $T_0(\delta)$. First we define the two events
\begin{align*}
    H_1(\epsilon) &= \min \left\{T \in \mathbb{N}^+: T - \sqrt{T} \geq T/(1+\epsilon)\right\},\\
    H_2(\epsilon) &= \min \left\{T \in \mathbb{N}^+: \Tilde{c}(\sqrt{T}) \in \left[\frac{\Delta^2}{8} - \epsilon, \frac{\Delta^2}{8} + \epsilon\right] \right\},
\end{align*}
which allows us to initially simplify the definition for $T_0(\delta)$ as follows where we denote $C^*_\epsilon = \frac{\Delta^2}{8} - \epsilon$
\begin{align}
    T_0(\delta) &= \min\left\{T \in \mathbb{N}^+: \sqrt{T} + \frac{\beta(T,\delta)}{\Tilde{c}(\sqrt{T})} \leq T \right\}\nonumber\\
    &\leq H_1(\epsilon) + \min\left\{T \in \mathbb{N}^+: \beta(T,\delta) \leq \frac{\Tilde{c}(\sqrt{T})T}{1+\epsilon}\right\}\nonumber\\
    &\leq H_1(\epsilon) + H_2(\epsilon) + \min\left\{T \in \mathbb{N}^+: \beta(T,\delta) \leq \frac{C^*_\epsilon T}{1+\epsilon}\right\}\nonumber\\
    &\leq H_1(\epsilon) + H_2(\epsilon)+ \min\left\{T \in \mathbb{N}^+: \log\left(T\gamma(K-1)/\delta\right)+ 8\log\log(T\gamma(K-1)/\delta) \leq \frac{C^*_\epsilon T}{1+\epsilon}\right\}\nonumber\\
    &\leq H_1(\epsilon) + H_2(\epsilon)+ \max \left(\frac{(1+\epsilon)^2}{C^*_\epsilon} \log\left(\frac{e(1+\epsilon)^2 \gamma(K-1)}{C^*_\epsilon \delta} \log \left(\frac{(1+\epsilon)^2 \gamma(K-1)}{C^*_\epsilon \delta}\right)\right), \frac{\delta}{\gamma(K-1)} \exp\left(g(\epsilon/8)\right)\right)\label{eqn_t0_final}
\end{align}
Where for equation \eqref{eqn_t0_final} we have used Lemma \ref{lemma_helpful_garivier}. Hence, by plugging in the upper bound from \eqref{eqn_t0_final} into \eqref{eqn_1cp_asymp_proof_before_division}, and observing the dependence of the upper bound on $\delta$ we attain the asymptotic upper bound stated in the theorem. Hence CPI has asymptotically optimal expected sample complexity.

% Hence by looking at the the dependence of $\delta$ on $T_0(\delta)$ in equation \eqref{eqn_t0_final}, as well as in our upper bound for the expected sample complexity in Proposition \ref{prop_complexity_UB}, we can construct the stated asymptotic upper bound for the expected sample complexity. 
\qed

\subsection*{Helpful Lemma}
We state the following helpful lemma from \cite{garivier2018thresholdingbanditdoserangingimpact}. For $0 < y \leq 1/e$, let $g$ be the function defined by 
\begin{equation*}
    g(y) := \frac{1}{y} \log\left(\frac{e}{y}\log\left(\frac{1}{y}\right)\right).
\end{equation*}
From this definition they state the following inequality in Lemma 10 \cite{garivier2018thresholdingbanditdoserangingimpact}.
\begin{lemma}\label{lemma_helpful_garivier}
    Let $A,B > 0$, then for all $\epsilon \in (0,1)$ such that $(1+\epsilon)/A < e$ and $B/\epsilon > e$, for all $x \geq \max \left(g(A/(1+\epsilon),\exp(g(\epsilon/B)))\right)$ we have
    \begin{equation*}
        \log(x) + B\log \log (x) \leq Ax
    \end{equation*}
\end{lemma}

%%%%%%%%%%%%%%%%%%%%%%%%%%%%%%%%%%%%%%%%%%%%%%%%%%%%%%%%%%%%%%%%%%%%%%%%%%%%%%%
%%%%%%%%%%%%%%%%%%%%%%%%%%%%%%%%%%%%%%%%%%%%%%%%%%%%%%%%%%%%%%%%%%%%%%%%%%%%%%%
%%%%%%%%%%%%%%%%%%%%%%%%%%%%%%%%%%%%%%%%%%%%%%%%%%%%%%%%%%%%%%%%%%%%%%%%%%%%%%%
%%%%%%%%%%%%%%%%%%%%%%%%%%%%%%%%%%%%%%%%%%%%%%%%%%%%%%%%%%%%%%%%%%%%%%%%%%%%%%%
\newpage
\section*{Proofs for Section \ref{section:multiple_cp}}
\section{Proof of Theorem \ref{theorem_exact_m_lb}}\label{app_proof_theorem_exact_m_lb}
We first note that the proof of the first inequality \eqref{eqn_theorem_exact_m_lb1} is similar to Theorem \ref{theorem_1cp_general} in Appendix \ref{app_proof_theorem_1cp_general}, except we now restrict to the set of environments with exactly $N$ change points (i.e. $V_{K,N}$). In order to prove the second inequality \eqref{eqn_theorem_exact_m_lb2} we simplify and only consider the set of alternative environments in which we shift one of the change points in $v$ by one to the left or the right.

Consider the following two sets of alternatives $\{v'_j\}_{j=1}^N, \{v''_j\}_{j=1}^N \subset V_{K,N}$ and we shift only the left/right mean reward adjacent to each of the change points in $v$. In particular, we define
\begin{align*}
    \mu_{v',i} &= 
    \begin{cases}
        \mu_{v,i+1} \quad \text{for} \quad i=\underline{x}^*_{v,j}\\
        \mu_{v,i} \quad \text{otherwise}
    \end{cases}\\
    \mu_{v'',i} &= 
    \begin{cases}
        \mu_{v,i-1} \quad \text{for} \quad i=\underline{x}^*_{v,j}+1\\
        \mu_{v,i} \quad \text{otherwise}
    \end{cases}
\end{align*}

Now, by denoting $V'' =  \{v'_j\}_{j=1}^N \cup \{v''_j\}_{j=1}^N \subset V^{alt}_{K,N}(\underline{x}^*_v)$. We have that from the definition of $c^*_2(v)^{-1}$ that

\begin{align}
    c^*_2(v)^{-1} &=  \sup_{\alpha \in \mathcal{P}_{K}}\inf_{v' \in V^{alt}_{K,N}(\underline{x}^*_v)} \sum_{i=1}^K \alpha_i D(v_i,v_i')\nonumber\\
    &\leq \sup_{\alpha \in \mathcal{P}_{K}}\inf_{v' \in V''} \sum_{i=1}^K \alpha_i D(v_i,v_i')\nonumber\\
    &=\sup_{\alpha \in \mathcal{P}_K} \min_{j} \left\{ \min \left\{\frac{\Delta_j^2}{2\sigma^2} \alpha_{\underline{x}^*_{v,j}},\frac{\Delta_j^2}{2\sigma^2} \alpha_{\underline{x}^*_{v,j}+1}\right\} \right\}\label{eqn_sup_subset_bound}
\end{align}

Hence, by substituting \eqref{eqn_sup_subset_bound} into equation \eqref{eqn_theorem_exact_m_lb1} of Theorem \ref{theorem_exact_m_lb} we get the following.
\begin{align}
    \log\left(\frac{1}{4\delta}\right)
 &\leq  \mathbb{E}_{\pi,v}[\tau] \sup_{\alpha \in \mathcal{P}_K} \min_{j} \left\{ \min \left\{\frac{\Delta_j^2}{2\sigma^2} \alpha_{\underline{x}^*_{v,j}},\frac{\Delta_j^2}{2\sigma^2} \alpha_{\underline{x}^*_{v,j}+1}\right\} \right\}\nonumber\\
 &=\mathbb{E}_{\pi,v}[\tau] \frac{1}{4\sigma^2 \left(\sum_{i=1}^{m} \frac{1}{\Delta_i^2}\right)}\nonumber\\
 \Longrightarrow \mathbb{E}_{\pi,v}[\tau] &\geq 4\sigma^2 \log\left(\frac{1}{4\delta}\right) \left(\sum_{i=1}^{m} \frac{1}{\Delta_i^2}\right)\nonumber
\end{align}
As required. \qed

%%%%%%%%%%%%%%%%%%%%%%%%%%%%%%%%%%%%%%%%%%%%%%%%%%%%%%%%%%%%%%%%%%%%%%%%%%%%%%%
%%%%%%%%%%%%%%%%%%%%%%%%%%%%%%%%%%%%%%%%%%%%%%%%%%%%%%%%%%%%%%%%%%%%%%%%%%%%%%%
%%%%%%%%%%%%%%%%%%%%%%%%%%%%%%%%%%%%%%%%%%%%%%%%%%%%%%%%%%%%%%%%%%%%%%%%%%%%%%%
%%%%%%%%%%%%%%%%%%%%%%%%%%%%%%%%%%%%%%%%%%%%%%%%%%%%%%%%%%%%%%%%%%%%%%%%%%%%%%%
\section{Proof of Theorem \ref{theorem_any_m_lb_specific_case}}\label{app_proof_theorem_any_m_lb_specific_case}

Consider any alternative environment $v' \in V_{K,m}$ where $m\geq N$ such that the change points in $v$ are not a subset of the change points in $v'$, namely $\underline{x}^*_v \not\subseteq \underline{x}^*_{v'}$. Then, since $\pi$ is an \textbf{Any-$(N,\delta)$} policy and equation \eqref{defn_any1_objective}, we have that 
$\underline{x}^*_v \not\subseteq \underline{x}^*_{v'}$ implies that $\hat{\underline{x}}_\tau = \underline{x}^*_v$ is a failure event in environment $v'$ occurring with small probability less than
\begin{equation} \label{eqn_BH_M_equals_m_1}
    \delta \geq \mathbb{P}_{\pi,v'}(\hat{\underline{x}}_\tau = \underline{x}^*_v).
\end{equation}
Furthermore, again due to equation \eqref{defn_any1_objective}, we have that under environment $v$ we have $\hat{\underline{x}}_\tau \neq \underline{x}^*_v$ occurring with small probability less than 
\begin{equation} \label{eqn_BH_M_equals_m_2}
    \delta \geq \mathbb{P}_{\pi,v}(\hat{\underline{x}}_\tau \neq \underline{x}^*_v)
\end{equation}
Combining equation \eqref{eqn_BH_M_equals_m_1} and \eqref{eqn_BH_M_equals_m_2} gives the following, where we apply the Bretagnolle-Huber Inequality \citep[Theorem 14.2][]{BanditAlgosBook} in equation \eqref{eqn_BH_M_equals_m_3}.
\begin{align}
    2\delta &\geq \mathbb{P}_{\pi,v}(\hat{\underline{x}}_\tau \neq \underline{x}^*_v) + \mathbb{P}_{\pi,v'}(\hat{\underline{x}}_\tau = \underline{x}^*_v) \nonumber\\
    &\geq \frac{1}{2} \exp\left(-D(\mathbb{P}_{\pi,v},\mathbb{P}_{\pi,v'})\right) \label{eqn_BH_M_equals_m_3}\\
    &\geq \frac{1}{2} \exp\left(-\sum_{i=1}^K 
 \mathbb{E}_{v,\pi}[T_i(\tau)]D\left(v(i),v'(i)\right)) \right) \nonumber\\
 \Longrightarrow \log\left(\frac{1}{4\delta}\right) &\leq \sum_{i=1}^K 
 \mathbb{E}_{v,\pi}[T_i(\tau)]D\left(v(i),v'(i)\right))\label{eqn_BH_M_equals_m_4}
\end{align}
Now, consider a set of $N$ alternative environments $\{v'_j\}_{j=1}^N \subset V_{K,m}$ where $m>N$. Let environment $v'_j$ have mean rewards equal to $v$ everywhere except for actions $a_j:=\underline{x}^*_{v,j}$ and $b_j:= \underline{x}^*_{v,j}+1$ such that there is no longer a change point there. In particular, for some $\ell_j \in [\mu_{v,a_j},\mu_{v,b_j}]$ we set
\begin{equation*}
    \mu_{v',i} = 
    \begin{cases}
         \ell_j \quad \text{when} \quad i = a_j \quad \text{or} \quad i=b_j\\
        \mu_{v,i}\quad \text{otherwise}
    \end{cases}
\end{equation*}
Now, since equation \eqref{eqn_BH_M_equals_m_4} holds for any of the alternative environments in the set $\{v'_j\}_{j=1}^m$, we have that the following set of inequalities.
\begin{align}
    \log\left(\frac{1}{4\delta}\right) &\leq \min_{j} \left\{ \inf_{\ell_j} \sum_{i=1}^K 
 \mathbb{E}_{v,\pi}[T_i(\tau)]D\left(v(i),v_j'(i)\right)\right\}\nonumber\\
 &\leq \mathbb{E}_{\pi,v}[\tau] \sup_{\alpha \in \mathcal{P}_K} \min_{j} \left\{ \inf_{\ell_j} \sum_{i=1}^K 
 \alpha_i D\left(v(i),v_j'(i)\right))\right\}\nonumber\\
 &=  \mathbb{E}_{\pi,v}[\tau] \sup_{\alpha \in \mathcal{P}_K} \min_{j} \left\{ \frac{\Delta_j^2}{2\sigma^2} \frac{\alpha_{\underline{x}^*_{v,j}}\alpha_{\underline{x}^*_{v,j}+1}}{\alpha_{\underline{x}^*_{v,j}}+\alpha_{\underline{x}^*_{v,j}+1}}\right\} \label{eqn_M_equals_m_sup}\\
 &=\mathbb{E}_{\pi,v}[\tau] \frac{1}{8\sigma^2 \left(\sum_{i=1}^{m} \frac{1}{\Delta_i^2}\right)}\nonumber\\
 \Longrightarrow \mathbb{E}_{\pi,v}[\tau] &\geq 8\sigma^2 \log\left(\frac{1}{4\delta}\right) \left(\sum_{i=1}^{m} \frac{1}{\Delta_i^2}\right)\nonumber
\end{align}
Where in equation \eqref{eqn_M_equals_m_sup} we have chosen the minimizing choice for $l_j$ in each environment $v_j'$. Furthermore, we can show that the choice for $\alpha$ which attains the supremum in equation \eqref{eqn_M_equals_m_sup} is
\begin{equation*}
    \alpha^*_{j} = \alpha^*_{j+1}=\frac{\frac{1}{\Delta_j^2}}{2\sum_{i=1}^{N} \frac{1}{\Delta_i^2}}
\end{equation*}
for $j \in \{x^*_{v,1},\dots,x^*_{v,N}\}$, and zero for all other $\alpha^*_j$ values.
for $j \in \{1,\dots,m\}$, and zero for all other $\alpha^*$ values. The proof is then complete. \qed

%%%%%%%%%%%%%%%%%%%%%%%%%%%%%%%%%%%%%%%%%%%%%%%%%%%%%%%%%%%%%%%%%%%%%%%%%%%%%%%
%%%%%%%%%%%%%%%%%%%%%%%%%%%%%%%%%%%%%%%%%%%%%%%%%%%%%%%%%%%%%%%%%%%%%%%%%%%%%%%
%%%%%%%%%%%%%%%%%%%%%%%%%%%%%%%%%%%%%%%%%%%%%%%%%%%%%%%%%%%%%%%%%%%%%%%%%%%%%%%
%%%%%%%%%%%%%%%%%%%%%%%%%%%%%%%%%%%%%%%%%%%%%%%%%%%%%%%%%%%%%%%%%%%%%%%%%%%%%%%
\newpage
\section{Proof of Theorem \ref{theorem_any_m_lb_general_case}}\label{app_proof_theorem_any_m_lb_general_case}

We will use a similar approach to the proof of Theorem \ref{theorem_any_m_lb_specific_case} in Appendix \ref{app_proof_theorem_any_m_lb_specific_case}, except we will firstly use a slightly different change of measure argument instead of the Bretagnolle-Huber inequality. Recall from Lemma 1 in \cite{Garivier_chain_rule_trick} we have that
\begin{equation}\label{eqn_transportation_lemma}
    D(\mathbb{P}_1,\mathbb{P}_2) \geq kl(\mathbb{P}_1(A), \mathbb{P}_2(A)) \geq 0
\end{equation}
for some event $A$. Where $kl(x,y)$ is the KL-divergence between two Bernoulli distributions with parameters $x$ and $y$. Recall also that
\begin{equation}\label{eqn_useful_kl_trick}
    kl(x,y) \geq x\log(1/y) - \log(2) 
\end{equation}

Let $v \in V_{K,m}$ where $m>N$ and let $\pi$ be an Any-$(N,\delta)$ policy. Now, for $j\in [m]$ define \begin{equation}
    A_j = \{x^*_{(j)}\in \underline{\hat{x}}_\tau\}.\nonumber
\end{equation}
Furthermore, let environment $v'_j \in V_{K,m+1}$ be equal to $v$ everywhere except in $\{x^*_{v,(j)},x^*_{v,(j)+1}\}$ where the mean reward is equal to $l_j$, namely
\begin{equation}
    \mu_{v_j',i} = 
    \begin{cases}
        l_j \quad \text{for} \quad i\in \{x^*_{v,(j)},x^*_{v,(j)+1}\}\\
        \mu_{v,i} \quad \text{otherwise}
    \end{cases}\nonumber
\end{equation}
Using the transportation lemma \eqref{eqn_transportation_lemma}, we have for any $j\in [m]$ and for any $l_j \in \mathbb{R}$ that
\begin{align}
    kl(\mathbb{P}_{\pi,v}(A_j), \mathbb{P}_{\pi,v_j'}(A_j)) &\leq D(\mathbb{P}_{\pi,v},\mathbb{P}_{\pi,v_j'})\nonumber\\
    &= \sum_{i=1}^K 
 \mathbb{E}_{v,\pi}[T_i(\tau)]D\left(v(i),v_j'(i)\right) \label{eqn_about_to_apply_inf}
\end{align}
However, since \eqref{eqn_about_to_apply_inf} applies for any choice of $l_j$ in environment $v_j'$, we can take the following infimum over the RHS.

\begin{align}
    kl(\mathbb{P}_{\pi,v}(A_j), \mathbb{P}_{\pi,v_j'}(A_j)) &\leq \inf_{l_j \in \mathbb{R}}\sum_{i=1}^K 
 \mathbb{E}_{v,\pi}[T_i(\tau)]D\left(v(i),v_j'(i)\right)\nonumber\\
 &= \mathbb{E}_{\pi,v}\left[T_{x^*_{v,(j)}}(\tau) + T_{x^*_{v,(j)+1}}(\tau)\right] \frac{\Delta_{(j)}^2}{8\sigma^2} \label{eqn_about_to_apply_min}
\end{align}
Now, \eqref{eqn_about_to_apply_min} holds for any $j\in [m]$. Hence, if we define $k_j = kl(\mathbb{P}_{\pi,v}(A_j), \mathbb{P}_{\pi,v_j'}(A_j))$ and $t_j = \mathbb{E}_{\pi,v}\left[T_{x^*_{v,(j)}}(\tau) + T_{x^*_{v,(j)+1}}(\tau)\right]$, then from \eqref{eqn_about_to_apply_min} gives us
\begin{align}
    1 &\leq \min_{j\in[m]} \left\{\frac{t_j\Delta_{(j)}^2}{8\sigma^2k_j}\right\}\nonumber\\
    &\leq \sup_{t_j:\sum_{j=1}^m t_j \leq \mathbb{E}_{\pi,v}[\tau]} \min_{j\in[m]}\left\{\frac{t_j\Delta_{(j)}^2}{8\sigma^2k_j}\right\}\label{eqn_sup_sum_Et_step}\\
    &= \frac{\mathbb{E}_{\pi,v}[\tau]}{\sum_{j=1}^m \frac{8\sigma^2k_j}{\Delta_{(j)}^2}}\label{eqn_sup_sum_Et_step_solved}
\end{align}
Here \eqref{eqn_sup_sum_Et_step} comes from taking a supremum over the potential values for $t_j$, taking into account that their sum is upper bounded by $\mathbb{E}_{\pi,v}[\tau]$ and \eqref{eqn_sup_sum_Et_step_solved} comes from finding the explicit solution to this supremum (similar to other lower bound proofs in this paper). Now, we try to lower bound the denominator of \eqref{eqn_sup_sum_Et_step_solved}.
\begin{align}
    \sum_{j=1}^m \frac{8\sigma^2k_j}{\Delta_{(j)}^2} &= \sum_{j=1}^m \frac{8\sigma^2 \, kl(\mathbb{P}_{\pi,v}(A_j), \mathbb{P}_{\pi,v_j'}(A_j))}{\Delta_{(j)}^2}\label{eqn_from_jk_defn}\\
    &\geq\sum_{j=1}^m \frac{8\sigma^2} {\Delta_{(j)}^2} \left(\mathbb{P}_{\pi,v}(A_j)\log(1/\mathbb{P}_{\pi,v_j'}(A_j)) - \log(2) 
\right) \label{eqn_from_kl_trick}\\
&\geq\sum_{j=1}^m \frac{8\sigma^2} {\Delta_{(j)}^2} \left(\mathbb{P}_{\pi,v}(A_j)\log(1/\delta) - \log(2) 
\right) \label{eqn_from_pi_assumption}\\
&=\sum_{j=1}^m \frac{8\sigma^2} {\Delta_{(j)}^2} \mathbb{P}_{\pi,v}(A_j)\log(1/\delta) - \sum_{j=1}^m \frac{8\sigma^2} {\Delta_{(j)}^2}\log(2) 
 \label{eqn_expanding_brackets}
\end{align}
Here \eqref{eqn_from_jk_defn} comes from the definition of $k_j$, \eqref{eqn_from_kl_trick} comes from using the trick in \eqref{eqn_useful_kl_trick}, and \eqref{eqn_from_pi_assumption} comes from the Any-$(N,\delta)$ assumption on $\pi$.

Now, by defining $p_S = \mathbb{P}_{\pi,v}(\hat{x}_\tau = S)$, we note that 
$$\mathbb{P}_{\pi,v}(A_j) = \mathbb{P}_{\pi,v}(x^*_{(j)}\in \underline{\hat{x}}_\tau) = \sum_{S\subset[K]}p_S\mathbb{I}\{x^*_{(j)}\in S, |S|=N\} \geq \sum_{S \subset [K]}p_S \mathbb{I}\{x^*_{(j)}\in S, |S|=N, S \subset \underline{x}^*_v\} $$
Hence we can lower bound the first sum on the right hand side of \eqref{eqn_expanding_brackets} by a linear function of the variables $\{p_S : |S|=N, S \subset \underline{x}^*_v\}$, namely
\begin{equation}\label{eqn_writign_as_linear_sum}
    \sum_{j=1}^m \frac{8\sigma^2} {\Delta_{(j)}^2} \mathbb{P}_{\pi,v}(A_j)\log(1/\delta) \geq \sum_{j=1}^m \frac{8\sigma^2} {\Delta_{(j)}^2} \log(1/\delta)\sum_{S \subset [K]}p_S \mathbb{I}\{x^*_{(j)}\in S, |S|=N, S \subset \underline{x}^*_v\} .
\end{equation}
Since the right hand side of \eqref{eqn_writign_as_linear_sum} is a linear function of the $p_S$ variables and by the fact that $\pi$ is Any-$(N,\delta)$ we have $1-\delta \leq \sum p_S \mathbb{I}\{|S|=N, S \subset \underline{x}^*_v\} \leq 1$. Hence, by Bauer's Maximum Principle \citep{Bauer1958MinimalstellenVF_maximum_principle}, the right-hand-side of \eqref{eqn_writign_as_linear_sum} is minimized at an extrema of the set of potential $p_S$ variables. Namely $p_S = 1-\delta$ for some $S=S^* \subset \underline{x}^*_v$ and $p_S=0$ otherwise. In particular we see that a minimiser is with $S^* = \{x^*_{v,(1)}, \dots, x^*_{v,(N)}\}$ such that we get
\begin{align}
    \sum_{j=1}^m \frac{8\sigma^2} {\Delta_{(j)}^2} \mathbb{P}_{\pi,v}(A_j)\log(1/\delta) &\geq \min_{S \subset \underline{x}^*_v, |S|=N}\sum_{j=1}^m \frac{8\sigma^2} {\Delta_{(j)}^2} \log(1/\delta)\sum_{S \subset [K]}p_S \mathbb{I}\{x^*_{(j)}\in S, |S|=N, S \subset \underline{x}^*_v\}\nonumber\\
    &=(1-\delta)\sum_{j=1}^N \frac{8\sigma^2} {\Delta_{(j)}^2} \log(1/\delta).\label{eqn_from_maximum_principle}
\end{align}
The, by plugging \eqref{eqn_from_maximum_principle} back into \eqref{eqn_expanding_brackets}. Then plugging \eqref{eqn_expanding_brackets} back into \eqref{eqn_sup_sum_Et_step_solved} we get our lower bound.

%%%%%%%%%%%%%%%%%%%%%%%%%%%%%%%%%%%%%%%%%%%%%%%%%%%%%%%%%%%%%%%%%%%%%%%%%%%%%%%
%%%%%%%%%%%%%%%%%%%%%%%%%%%%%%%%%%%%%%%%%%%%%%%%%%%%%%%%%%%%%%%%%%%%%%%%%%%%%%%
%%%%%%%%%%%%%%%%%%%%%%%%%%%%%%%%%%%%%%%%%%%%%%%%%%%%%%%%%%%%%%%%%%%%%%%%%%%%%%%
%%%%%%%%%%%%%%%%%%%%%%%%%%%%%%%%%%%%%%%%%%%%%%%%%%%%%%%%%%%%%%%%%%%%%%%%%%%%%%%
\newpage
\section{Proof of Proposition \ref{prop_MCPI_ub_exp}}\label{app_proof_prop_MCPI_ub_exp}

We have the following concentration inequality as a consequence of Lemma 6 in \citep[][]{Degenne2019NonAsymptoticPE}.

\begin{lemma} \label{lemma_degenne_concentration}
    Define the event
    \begin{equation}\label{good_event_final_upper_bound}
        \mathcal{E}_t = \left\{ \forall s \leq t, \forall a \in [K] \quad |\hat{\mu}_{a}(s) - \mu_{a}|^2 \leq \frac{4\log(t) + 2\log(2\log(t)) +1}{T_a(s )} \right\}
    \end{equation}
    Then 
\begin{equation*}
    \forall t \geq 3, \quad \mathbb{P}_\mu (\mathcal{E}_t^c) \leq 2eK \frac{\log t}{t^{2}}, \quad \sum_{t=3}^{+\infty} \mathbb{P}_\mu (\mathcal{E}_t^c) \leq 2eK.
\end{equation*}
and 
\[
\sum_{t=3}^{+\infty} \mathbb{P}_\mu (\mathcal{E}_t^c) \leq 2eK \sum_{t=3}^{+\infty} \frac{\log t}{t^{2}} \leq 2eK \int_{x=1}^{+\infty} \frac{\log x}{x^{2}} dx = 2eK.
\]
\end{lemma}

Now, if we define 
\begin{equation*}
    r(t) = \sqrt{\frac{4\log(t) + 2\log(2\log(t)) + 1/2}{(t^{1/4}-K)_+}},
\end{equation*}

then we have that 
\begin{equation*}
    \mathcal{E}_t \Longrightarrow \mathcal{E}'_t = \bigg\{\forall s \in [t^{1/2},t], \forall a \in [K] \quad |\hat{\mu}_{a}(s) - \mu_{a}| \leq r(t) \bigg\}.
\end{equation*}
This is true since, for $s \in [t^{1/2},t]$, our forced exploration in MCPI means that $T_a(s) \geq (s^{1/2}-K)_+ \geq (t^{1/4}-K)_+$. We now state a useful Lemma which is a consequence of the definition of our simple estimator \eqref{eqn_estimator_simple_defn} and the event $\mathcal{E}'_t$.

\begin{lemma}
    Define 
    \begin{equation*}
        T_1'(v) = \min \left\{T \in \mathbb{N}^+ : r(T) < \frac{\Delta_{(N)}- \Delta_{(\ell)}}{4}\right\}.
    \end{equation*}
    Then, for $t\geq T_1'(v)$, we have that 
    \begin{equation*}
    \mathcal{E}_t' \Longrightarrow B_t = \bigg\{\forall s \in [t^{1/2},t], \hat{x}_s \in \{x_{(1)}, \dots, x_{(N)}\} \bigg\}.
\end{equation*}
\end{lemma}

Note that event $B_t$ implies that all tracking actions (see line 14 Algorithm \ref{algo_TandS2}) that are played in rounds $ s \in [t^{1/2},t]$ will be in the set
$$a_s \in \{x^*_{(1)}, x^*_{(1)}+1, \dots, x^*_{(N)}, x^*_{(N)}+1\}$$
We can now present the following Lemma.

\begin{lemma}
    The total number of tracking actions MCPI will play in rounds $s\in [t^{1/2},t]$ before stopping, under event $\mathcal{E}_t'$, is at most
    \begin{equation*}
        \sum_{j=1}^N \frac{8(t,\delta/N)}{(\Delta_{(j)}-2r(t))^2}.
    \end{equation*}
\end{lemma}
\begin{proof}
    Let us denote $n_j$ to be the number of tracking actions in which we have played action $j$ between rounds $t^{1/4}$ and $t$. Also let
$$N_j = \min\{n_{x^*_{(j)}},n_{x^*_{(j)}+1}\}$$
be the smallest number of tracking actions played by $x^*_{(j)}$ or $x^*_{(j)}+1$.

Now, let $s \in [t^{1/2},t]$. If we then define 
\begin{equation*}
    \Tilde{Z}_{j}(s) = \frac{T_{x^*_{(j)}}(s)T_{x^*_{(j)}+1}(s)}{2(T_{x^*_{(j)}}(s)+ T_{x^*_{(j)}+1}(s))} \hat{\Delta}_{x^*_{(j)}}^2(s).
\end{equation*}
Then since $T_{x^*_{(j)}}(s), T_{x^*_{(j)}+1}(s) \geq N_j$, we have

\begin{equation*}
    \Tilde{Z}_{j}(s) \geq \frac{N_j N_j}{2(N_j + N_j)} \hat{\Delta}_{x^*_{(j)}}^2(s).
\end{equation*}
Therefore, under our event $\mathcal{E}_t'$,
\begin{equation}\label{eqn_z_tilde_lower_b}
    \Tilde{Z}_{j}(s) \geq \frac{N_j}{4} (\Delta_{(j)}-2r(t))^2.
\end{equation}
Note, by the definition of our stopping time ( see \eqref{eqn_defn_stopping_time} and line 6 of Algorithm \ref{algo_TandS2}) and since $\beta(t,\delta/N) > \beta(s,\delta/N)$; when 
$$\Tilde{Z}_{j}(s) \geq \beta(t,\delta/N)$$
occurs, we will never play actions $x^*_{(j)}$ or $x^*_{(j)}+1$ when tracking again. Hence, from \eqref{eqn_z_tilde_lower_b}, if 
\begin{equation}\label{eqn_Nj_bound_individual}
    \frac{N_j}{4} (\Delta_(j)-2r(t))^2
\end{equation}
holds then we will never play actions $x^*_{(j)}$ or $x^*_{(j)}+1$ when tracking again. By isolating the $N_j$ values in \eqref{eqn_Nj_bound_individual} and summing through $j\in [N]$ (i.e. summing over the total number of tracking action we can mae before stopping), we get the restul in the lemma.

\end{proof}

Now, in rounds $s\in [1,t]$ there will at most $K\sqrt{t}$ forced exploration actions. Hence in rounds $s\in [t^{1/2},t]$ there will be at least $(t-2K\sqrt{t})_+$ tracking actions. We therefore get the following lemma.

\begin{lemma} \label{lemma_t1t0_useful}
    If $t>T_1(v)$ and $t>T_0(\delta) = \min\left\{T \in \mathbb{N}^+: T -2KT^{\frac{1}{2}}\geq \sum_{i=1}^N\frac{8\beta(T,\delta/N)}{(\Delta_{(i)}-2r(T))^2} \right\}$. Then $$\mathcal{E}_t' \Longrightarrow \tau\leq t$$
\end{lemma}

Now, putting everything together using Lemma \ref{lemma_t1t0_useful} and Lemma \ref{lemma_degenne_concentration} gives us the following high probability upper bound for the stopping time for MCPI.

\begin{theorem}\label{final_hp_uper_bound}
    Let $t>\max\{T_1(v),T_0(\delta)\}$ then 
    \begin{equation*}
        \mathbb{P}(\tau<t) \leq \mathbb{P}(\mathcal{E}_t^C) \leq \frac{2eK\log(t)}{t^2}
    \end{equation*}
\end{theorem}
Then, as a consequence of Theorem \ref{final_hp_uper_bound}
\begin{align*}
        \mathbb{E}[\tau] &= \sum_{t = 0}^\infty \mathbb{P}(\tau_\delta >t)\\
        & \leq T_0(\delta) + T_1 + \sum_{t = T_0(\delta) + T_1}^\infty \mathbb{P}(\tau_\delta >t)\\
        & \leq T_0(\delta) + T_1 + \sum_{t = T_0(\delta) + T_1}^\infty 2eK\frac{\log(t)}{t^{2}}\\
        & \leq T_0(\delta)+ T_1 + 2eK
    \end{align*}
    As required.\qed

%%%%%%%%%%%%%%%%%%%%%%%%%%%%%%%%%%%%%%%%%%%%%%%%%%%%%%%%%%%%%%%%%%%%%%%%%%%%%%%
%%%%%%%%%%%%%%%%%%%%%%%%%%%%%%%%%%%%%%%%%%%%%%%%%%%%%%%%%%%%%%%%%%%%%%%%%%%%%%%
%%%%%%%%%%%%%%%%%%%%%%%%%%%%%%%%%%%%%%%%%%%%%%%%%%%%%%%%%%%%%%%%%%%%%%%%%%%%%%%
%%%%%%%%%%%%%%%%%%%%%%%%%%%%%%%%%%%%%%%%%%%%%%%%%%%%%%%%%%%%%%%%%%%%%%%%%%%%%%%
\section{Proof of Theorem \ref{theorem_upper_mcp_asymptotic}}\label{app_proof_theorem_upper_mcp_asymptotic}

    Now, to prove the asymptotic upper bound on the expectation. Note that the final two terms of the non-asymptotic upper bound (Theorem \ref{prop_MCPI_ub_exp}) do not depend on $\delta$ and hence they will vanish when diving through by $\log(1/\delta)$. Hence, we focus on studying the first term $T_0(\delta)$. To do so, define the following, where $\epsilon_2,\epsilon>0$ are any small real number.
    \begin{align*}
        T_2(\epsilon_2) &= \min\left\{T \in \mathbb{N}^+: T -2KT^{\frac{1}{2}}\geq T(1-\epsilon_2) \right\}\\
        T_3(\epsilon_3) &= \min\left\{T \in \mathbb{N}^+: \sum_{i=1}^N\frac{1}{(\Delta_{(i)}-2r(T))^2} \leq \sum_{i=1}^N\frac{1}{\Delta_{(i)}^2} + \epsilon_3 \right\}
    \end{align*}
    We can then write
    \begin{align*}
        T_0(\delta) &\leq T_2(\epsilon_2) + T_3(\epsilon_3) + \min\left\{T \in \mathbb{N}^+: T(1-\epsilon_2)\geq 8\beta(T,\delta/N)\left(\sum_{i=1}^N\frac{1}{\Delta_{(i)}^2} + \epsilon_3\right) \right\}\\
        &= T_2(\epsilon_2) + T_3(\epsilon_3) + \min\left\{T \in \mathbb{N}^+: T(1-\epsilon_2)\geq \log(1/\delta)\frac{8\beta(T,\delta/N)}{\log(1/\delta)}\left(\sum_{i=1}^N\frac{1}{\Delta_{(i)}^2} + \epsilon_3\right) \right\}\\
    \end{align*}
    Hence, we have
    \begin{align*}
        \limsup_{\delta\rightarrow 0} \frac{\mathbb{E}[\tau]}{\log(1/\delta)} &\leq \frac{1}{\log(1/\delta)}\min\left\{T \in \mathbb{N}^+: T(1-\epsilon_2)\geq 8\log(1/\delta)\left(\sum_{i=1}^N\frac{1}{\Delta_{(i)}^2} + \epsilon_3\right) \right\}\\
        &= \frac{8\log(1/\delta)}{\log(1/\delta)(1-\epsilon_2)}\left(\sum_{i=1}^N\frac{1}{\Delta_{(i)}^2} + \epsilon_3\right)\\
        &= \frac{8}{(1-\epsilon_2)}\left(\sum_{i=1}^N\frac{1}{\Delta_{(i)}^2} + \epsilon_3\right)
    \end{align*}
    However, $\epsilon_2$ and $\epsilon_3$ were arbitrary hence we can send them to zero $\epsilon_2,\epsilon_3 \rightarrow0$ to attain the upper bound. \qed

\newpage
\section{Additional Discussion}
\subsection{Complexity Comparison With Best Arm Identification}

In order to compare the complexity of best arm identification and change point identification we construct a simple example in which there is a unique 'best arm' (with largest mean) and exactly one change point as follows. Suppose we are in an environment with K arms ($\sigma^2$-Gaussian) with mean rewards $(\mu, \mu, \dots, \mu, \mu+\Delta)$ for some $\mu\in \mathbb{R}$. From \citep{kaufmann_16a}, the complexity of the best arm identification problem in this environment will asymptotically be of order 
\begin{equation*}
    K\frac{\sigma^2}{\Delta^2}\log\left(\frac{1}{\delta}\right).
\end{equation*}
 
Now, from our Corollary \ref{theorem_exact_1cp_lb_explicit} the change point identification problem complexity of order

\begin{equation*}
    \frac{\sigma^2}{\Delta^2}\log\left(\frac{1}{\delta}\right).
\end{equation*}
 
We no longer have a linear dependence on K in the change point identification problem since we can use our piecewise constant structure and information from across our action space to confidently locate the change in mean. On the other hand, in the best arm identification problem, we have to sample each of the individual K arms sufficiently to confidently identify the arm with the highest mean reward.

%%%%%%%%%%%%%%%%%%%%%%%%%%%%%%%%%%%%%%%%%%%%%%%%%%%%%%%%%%%%%%%%%%%%%%%%%%%%%%%
%%%%%%%%%%%%%%%%%%%%%%%%%%%%%%%%%%%%%%%%%%%%%%%%%%%%%%%%%%%%%%%%%%%%%%%%%%%%%%%

\end{document}